%% file: main.tex
\documentclass[10pt,twocolumn,letterpaper]{article}

\usepackage[pagenumbers]{cvpr} %

\input{preamble}

\definecolor{cvprblue}{rgb}{0.21,0.49,0.74}
\usepackage[pagebackref,breaklinks,colorlinks,allcolors=cvprblue]{hyperref}

\title{Faster-GS: Analyzing and Improving Gaussian Splatting Optimization}

\author{
Florian Hahlbohm$^{1}$\qquad
Linus Franke$^{2}$\qquad
Martin Eisemann$^{1}$\qquad
Marcus Magnor$^{1}$\vspace{0.2em}\\
\small
\begin{tabular}{@{}c c@{}}
$^1$Computer Graphics Lab, TU Braunschweig, Germany &
$^2$Inria, Universit\'e C\^ote d'Azur, France \\
{\tt \{lastname\}@cg.tu-bs.de} &
{\tt \{firstname.lastname\}@inria.fr}
\end{tabular}\\
\vspace{0.2em}
\small \url{https://fhahlbohm.github.io/faster-gaussian-splatting}
}

\begin{document}

\maketitle
\input{figures/0_teaser}
\input{sections/0_abstract}
\input{sections/1_introduction}
\input{sections/2_related_work}
\input{sections/3_method}
\input{sections/4_evaluation}
\input{sections/5_discussion}
\input{sections/6_conclusion}
{
    \small
    \bibliographystyle{ieeenat_fullname}
    \bibliography{main}
}
\appendix
\input{sections/X_suppl}

\end{document}

%% file: preamble.tex
\newcommand{\TODO}[1]{\textbf{\color{red}[TODO: #1]}}
\renewcommand{\TODO}[1]{}

\usepackage{microtype}

\renewcommand{\paragraph}[1]{\vspace{.5em}\noindent\textbf{#1.}}

\usepackage{nicefrac}
\usepackage{multicol, multirow}
\usepackage[table]{xcolor}
\definecolor{1st}{RGB}{102,194,164} %
\definecolor{2nd}{RGB}{178,226,226} %
\definecolor{3rd}{RGB}{217,248,241} %
\definecolor{1stText}{RGB}{57,146,116} %
\newcommand\tstrut{\rule{0pt}{2.4ex}}
\newcommand\bstrut{\rule[-1.0ex]{0pt}{0pt}}

\usepackage{dsfont}
\usepackage{enumitem}

\usepackage{cuted}

%% file: figures/0_teaser.tex
\begin{strip}
\centering
\vspace{-1.5cm}
\includegraphics[width=1.0\linewidth]{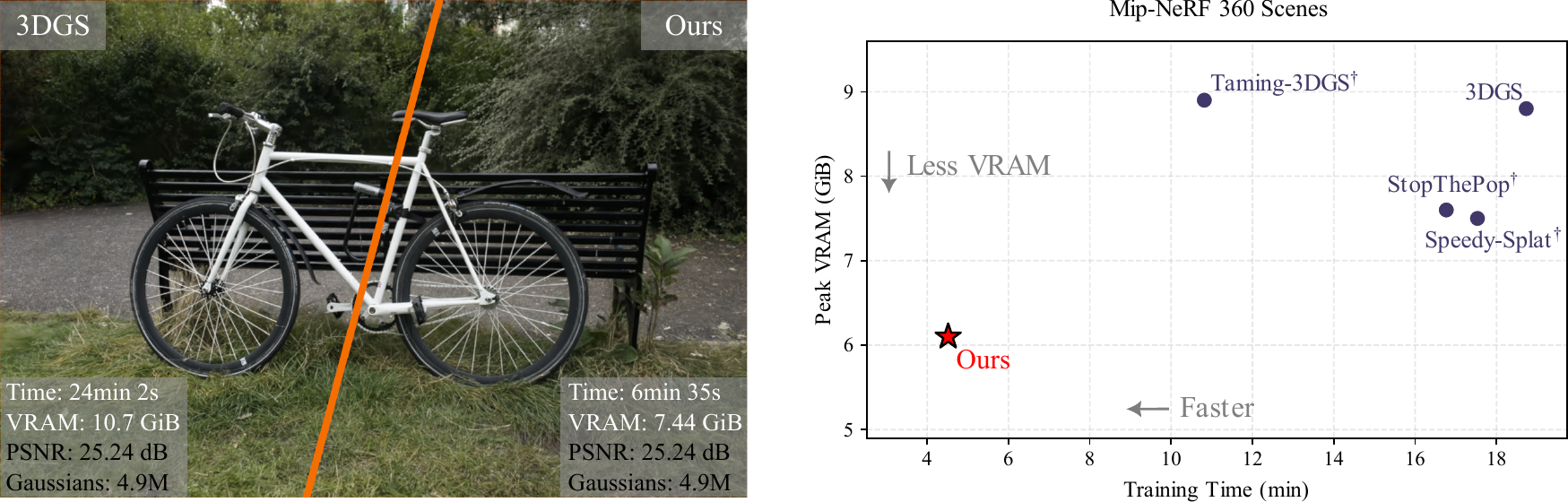}
\vspace{-0.3cm}
\captionsetup{type=figure}
\captionof{figure}{%
Our method, \textit{Faster-GS}, substantially accelerates training and reduces GPU memory (VRAM) usage compared to the original 3DGS algorithm~\cite{kerbl3Dgaussians} without altering quality or number of Gaussians (left). Averaged over all Mip-NeRF360 scenes on an RTX 4090 GPU (right), we train 4.1\texttimes\ faster with 30\% less VRAM than 3DGS. On the Deep Blending dataset (not depicted), the speedup is more than 5.2\texttimes. We also outperform improved implementations from prior works~\cite{taming3dgs, radl2024stopthepop, hanson2025speedysplat}. $^\dagger$Baseline modified to keep quality$\slash$\#Gaussians unchanged.
}\label{fig:teaser}
\end{strip}

%% file: sections/0_abstract.tex
\begin{abstract}
Recent advances in 3D Gaussian Splatting (3DGS) have focused on accelerating optimization while preserving reconstruction quality.
However, many proposed methods entangle implementation-level improvements with fundamental algorithmic modifications or trade performance for fidelity, leading to a fragmented research landscape that complicates fair comparison.
In this work, we consolidate and evaluate the most effective and broadly applicable strategies from prior 3DGS research and augment them with several novel optimizations.
We further investigate underexplored aspects of the framework, including numerical stability, Gaussian truncation, and gradient approximation.
The resulting system, Faster-GS, provides a rigorously optimized algorithm that we evaluate across a comprehensive suite of benchmarks.
Our experiments demonstrate that Faster-GS achieves up to 5\texttimes\ faster training while maintaining visual quality, establishing a new cost-effective and resource efficient baseline for 3DGS optimization.
Furthermore, we demonstrate that optimizations can be applied to 4D Gaussian reconstruction, leading to efficient non-rigid scene optimization.
\end{abstract}

%% file: sections/1_introduction.tex
\section{Introduction}
\label{sec:intro}

In their seminal work on \emph{3D Gaussian Splatting (3DGS)}~\cite{kerbl3Dgaussians}, 
Kerbl~\etal introduce a scene representation and novel view synthesis framework 
that unifies the strengths of classical point-based rendering~\cite{zwicker2001ewasplat} 
with gradient-based optimization techniques from differentiable volumetric 
rendering~\cite{mildenhall2020nerf}.

Owing to its remarkable combination of visual fidelity and real-time performance, 
3DGS rapidly became the dominant approach for novel view synthesis and inspired a broad 
range of subsequent research in computer vision and computer graphics~\cite{xiang2025structured,
liu2024humangaussian,matsuki2024gaussian,tosi2024nerfs}, as well as interest in 
domains such as digital film production~\cite{philip2025detail}.
The extensive adoption underscores the importance and impact of cost-effective reconstructions, the central theme of this paper.

The sustained momentum of this research area can partially be attributed to the rapid emergence 
of numerous extensions addressing specific aspects of the original formulation, including 
anti-aliasing~\cite{yu2024mip}, densification~\cite{kheradmand243dgsmcmc,
rotabulo2025revising}, compression~\cite{3DGSzip2024}, rendering approximations~\cite{radl2024stopthepop,
hahlbohm2025htgs}, and improvements in inference speed and scalability~\cite{schuetz2025splatshop,
taming3dgs,feng2025flashgs,kerbl2024hierarchical, liao2025tensorcoregs}.
Notably, recent performance-oriented variants 
achieve high-quality reconstructions within minutes, even on consumer-grade hardware, further 
accelerating research and experimentation in this field. However, the rapid pace of progress 
also poses a practical challenge: the continuous influx of improvements often outpaces the 
ability to integrate and evaluate them cohesively when developing new methods.

Our work is motivated by two main observations:
First, recent research on 3D Gaussian Splatting (3DGS) has led to a fragmented landscape 
of extensions and optimizations, making it increasingly difficult to assess the upper bound 
of achievable performance when integrating all available net-positive contributions.
This issue is particularly evident for \emph{training performance}, where the goal is to 
reduce optimization time while maintaining the reconstruction quality of the original 
3DGS method~\cite{kerbl3Dgaussians}. Advances in this domain are often tightly integrated with 
more fundamental modifications to the algorithm or underlying representation~\cite{radl2024stopthepop,hahlbohm2025htgs}. Other improvements, in contrast, have been developed primarily for inference (i.e., novel view synthesis rendering)~\cite{schuetz2025splatshop,feng2025flashgs}, 
although parts of these techniques could, in principle, be adapted for training, which we demonstrate in this work.

Second, while community-driven frameworks such as \textit{gsplat}~\cite{ye2025gsplat} have made significant progress toward extensible and modular implementations that facilitate fundamental modifications to the algorithm~\cite{huang20242d,wu20253dgut}, this hinders the integration of performance optimization targeting the original pipeline~\cite{zwicker2001ewasplat,kerbl3Dgaussians}, revealing a gap in the research landscape.

Our work aims to provide a solution to the core problem implied by these observations, \ie, the lack of an updated 3DGS baseline regarding training performance.
Thus, in this paper, we both survey the 3DGS follow-up works for performance improvements as well as integrate novel improvements. 
We evaluate their efficiency and integrate them into a new, optimized 3D Gaussian Splatting framework (\cf \cref{fig:teaser}).
Specifically, we observe that most recent approaches aim to \textbf{reduce memory accesses} on the GPU through fewer duplicated Gaussians during differentiable rasterization~\cite{radl2024stopthepop,hanson2025speedysplat,wang2024adrgaussian}, reduced memory accesses during sorting~\cite{schuetz2025splatshop}, or memory-efficient thread-to-workload setups~\cite{taming3dgs,feng2025flashgs}, all with varying degrees of effectiveness.
Additionally, several works report performance gains through \emph{kernel fusion}, 
where multiple computational stages are combined into a single pass~\cite{hahlbohm2025htgs,franke2025vrsplatting,taming3dgs}.

Beyond \textbf{integrating}, comparing, and evaluating these established techniques, we reveal and introduce methods to further decrease memory access costs by leveraging \textbf{memory coalescence}, which improves cache locality 
and bandwidth utilization. 
We apply all optimizations to the training schedule of 3DGS and showcase a speedup of up to $5\times$ in training speeds, resulting in an average reconstruction time of 163~seconds for the 3DGS benchmark~\cite{kerbl3Dgaussians} with full quality and all Gaussians.
We specifically exclude lower-precision and hardware-level optimizations~\cite{liao2025tensorcoregs, li2025gaurast}, pruning strategies~\cite{hanson2025speedysplat}, dense Gaussian initialization~\cite{kotovenko2025edgs}, or feed-forward pipelines~\cite{jiang2025anysplat, charatan2024pixelsplat} as they fundamentally change results, whereas our aim is to retain compatibility with the original widely-used CUDA-based differentiable rasterization pipeline~\cite{kerbl3Dgaussians}.

We further showcase the effectiveness of this work and its potential impact on future 3DGS-based research by extending our implementation to multi-dimensional Gaussians (4D) for dynamic scene reconstruction based on the work by Yang~\etal~\cite{yang20244dgaussians}.
We make the following key contributions:
\begin{itemize}
\item A comprehensive survey, discussion, and evaluation of 3DGS training optimization strategies, providing a comparative analysis of previously published improvements.
\item Novel performance optimizations that exploit \textit{memory coalescence} and \textit{fuse gradient computations and parameter updates}, significantly accelerating training without compromising reconstruction quality.
\item The introduction of \textit{Faster-GS}, an integrated and state-of-the-art 3DGS training pipeline that consolidates all effective techniques for real-time performance.
\end{itemize}
Our full implementation, including code$\slash$scripts for the presented experiments, are available on our project page.
Apart from a plug-and-play solution for existing 3DGS methods, it provides a testbed designed to facilitate future research 
and fair comparison within this rapidly evolving domain.

%% file: sections/2_related_work.tex
\section{Preliminaries and Related Work}
\label{sec:related_work}

\subsection{3D Gaussian Splatting (3DGS)}
\label{ssec:3dgs_recap}
To model a scene, 3DGS~\cite{kerbl3Dgaussians} uses a set of unnormalized 3D Gaussian point primitives, each of which is defined by its 3D mean $\mu$, anisotropic 3D covariance $\Sigma$, and a scalar opacity $o\in(0, 1)$.
Additionally, each primitive is equipped with a set of spherical harmonics (SH) coefficients to represent view-dependent color changes.
In practice, 3DGS uses coefficients up to degree three $\hat{c}\in\mathbb{R}^{3\times16}$ that can be evaluated for a given viewing direction to obtain an RGB color $c$.

\paragraph{Rendering}
To render a Gaussian, 3DGS first transforms relevant Gaussian parameters to camera space and employs splatting~\cite{zwicker2001ewasplat} to obtain a 2D Gaussian in screen space.
\begin{equation}\label{eq:mean_projection}
\mu_\text{2D} = \begin{pmatrix}
\frac{f_x \hat{\mu}_x}{\hat{\mu}_z} + c_x\\
\frac{f_y \hat{\mu}_y}{\hat{\mu}_z} + c_y
\end{pmatrix}\!, \; \text{with} \; \hat{\mu} = W(\mu_x, \mu_y, \mu_z, 1)^{\!\top}
\end{equation}\vspace{-2mm}
\begin{equation}\label{eq:cov_projection}
\Sigma_\text{2D} = J W_{\!1:3, 1:3} \Sigma W_{\!1:3, 1:3}^{\!\top} J^{\!\top}, \; J = \begin{pmatrix}
\frac{f_x}{\hat{\mu}_z} & 0 & -\frac{f_x \hat{\mu}_x}{\hat{\mu}_z^2}\\
0 &\frac{f_y}{\hat{\mu}_z}& -\frac{f_y \hat{\mu}_y}{\hat{\mu}_z^2}%
\end{pmatrix}
\end{equation}
where $W\in\mathbb{R}^{4\times4}$ the transformation from world to camera space, $J\in\mathbb{R}^{2\times3}$ the affine approximation of the perspective projection, and $f_x, f_y, c_x, c_y$ the intrinsic camera parameters, \ie, focal lengths and optical center in screen coordinates.
The result is a 2D Gaussian that can, after truncation (Kerbl~\etal~\cite{kerbl3Dgaussians} use $\approx\!3.33\sigma$), be rasterized and evaluated at any pixel position $x\in\mathbb{R}^{2\times1}$ efficiently, yielding a transparency value $\alpha$ for blending:
\begin{equation}\label{eq:alpha_computation}
\alpha = o \cdot e^{-\frac{1}{2}(x - \mu_\text{2D})^{\!\top}\Sigma^{-1}_\text{2D}(x - \mu_\text{2D})}.
\end{equation}
Pixel colors are computed by alpha-blending all $N$ fragments in each pixel in front-to-back order:
\begin{equation}\label{eq:alpha_blending}
C = \sum_{i=1}^{N} \alpha_i  T_i c_i + T_N c_\text{bg} \; \text{with} \quad T_i = \prod_{j=1}^{i-1} (1 - \alpha_j),
\end{equation}
where $c_\text{bg}$ is an arbitrary background color.
As an approximation to exact but prohibitively slow per-pixel sorting~\cite{radl2024stopthepop}, 3DGS employs approximate, object-level sorting, where 3D Gaussians are sorted based on $\hat{\mu}_z$ prior to rasterization.

\paragraph{Activation Functions}
To ensure stable, gradient-based optimization, 3DGS employs activation functions for all parameters except $\mu$.
Opacity is constrained to lie in $(0, 1)$ by applying a Sigmoid activation and the view-dependent color $c$ is clipped to be positive.
To ensure $\Sigma$ is a valid covariance, \ie, positive semi-definite, 3DGS optimizes separate scaling and rotation components from which $\Sigma$ is computed during rendering using $\Sigma\!=\!R S S^{\!\top} R^{\!\top}$, where $R$ is a 3D rotation matrix and $S$ is a diagonal scaling matrix.
$R$ is optimized as a quaternion that is normalized before rendering.
Furthermore, so that the scales of the 3D Gaussian remain positive, the exponential function is used as an activation prior to rendering.
Mathematically, this means $S$ stores the standard deviations of the 3D Gaussian along the principal axes defined by $R$, \ie, the three eigenvectors of $\Sigma$.

\paragraph{Optimization Schedule and Hyperparameters}
Starting from a sparse point cloud or random initialization, Kerbl~\etal~\cite{kerbl3Dgaussians} introduce adaptive density control (ADC), a set of carefully tuned heuristics for cloning, splitting, and pruning Gaussians during optimization.
Specifically, 3DGS tracks the magnitude of the $\mu_\text{2D}$ gradients during training for all Gaussians and clones or splits Gaussians at regular intervals.
Note that this growing set of Gaussians causes memory fragmentation, slowing down optimization in later iterations.

The training schedule and hyperparameters Kerbl~\etal provided with the initial code release have mostly remained unchanged in follow-up work.
Models are trained for 30000 iterations, in each of which an image is rendered from a randomly sampled training viewpoint and compared with the ground truth image, with losses as a combination of L1 and D-SSIM and parameter updates with Adam~\cite{kingma2014adam}.
For a full breakdown of the schedule, see \cref{sec:training_details}.

An important change that has been integrated into the official 3DGS codebase of Kerbl~\etal~\cite{kerbl3Dgaussians} since the initial release is a change in the opacity learning rate, which was halved from 0.05 to 0.025 following Mallick~\etal~\cite{taming3dgs}.
This change affects the number of Gaussians created during optimization, with the new, lower learning rate leading to slightly cleaner reconstructions with fewer Gaussians.

\subsection{Improvements and Follow-up Works}
\paragraph{Rendering}
At the heart of the original 3DGS algorithm and implementation by Kerbl~\etal~\cite{kerbl3Dgaussians} is a tile-based, differentiable software rasterizer implemented in C++$\slash$CUDA.
To reduce overhead, extensions for computing tight bounding boxes and exact splat$\slash$tile intersections have been proposed~\cite{radl2024stopthepop, wang2024adrgaussian, hanson2025speedysplat}.
A major effort has also been on reducing approximations and artifacts in 3DGS rendering by moving to ray-based evaluation schemes~\cite{moenne2024gaussiantracer, yu2024gof, hahlbohm2025htgs, talegaonkar2025vol3dgs} or proper volumetric rendering~\cite{condor2025dontsplat, blanc2025raygauss, blanc2025raygaussx, celarek2025does3dgsneedvolrend} to avoid distortion artifacts tied to the original splatting approach~\cite{huang2024erroranalysis}, improving blending accuracy at the pixel level~\cite{radl2024stopthepop, hahlbohm2025htgs, kv2025stochasticsplats, hou2025sortfree}, and suitable anti-aliasing strategies~\cite{yu2024mip, steiner2025aaags}.
Other works focus on improving performance specifically during inference, \eg, to speed up rendering when the number of primitives is high by devising optimized pipelines for sorting Gaussians~\cite{schuetz2025splatshop}, or improving the underlying data access patterns and control flow~\cite{feng2025flashgs, gui2024balanced3dgs}.
Complementary to these approaches, several works target architecture-level optimizations and use hardware acceleration~\cite{liao2025tensorcoregs, yang2025gsacc, li2025gaurast}, which trade configurability and numerical precision for performance.
Recent work also investigates efficient rendering on HMDs through foveated rendering~\cite{tu2025vrsplat, franke2025vrsplatting}.

\paragraph{Optimization}
During optimization, subsequent works improve the original algorithm by analyzing and enhancing the underlying densification heuristics~\cite{rotabulo2025revising, ye2024absgs, grubert2025improvedadc, fang2025minisplatting}, distributing Gaussians based on anchor points~\cite{lu2024scaffoldgs}, integrating probabilistic models~\cite{kheradmand243dgsmcmc}, or fully replacing densification in favor of dense initialization~\cite{kotovenko2025edgs}.
An equally important aspect of optimization is avoiding excessive growth in the number of Gaussians, \eg, by pruning obsolete Gaussians~\cite{girish2024eagles, niemeyer2024radsplat} or by reducing the number created during optimization indirectly by repeatedly reducing the opacity of each Gaussian~\cite{radl2024stopthepop}.
Other work addresses challenges commonly associated with real-world data, \eg, camera lens distortions~\cite{wu20253dgut}, dynamic distractors~\cite{sabour2025spotlesssplats}, and textureless regions and lighting variations, which can be resolved through depth regularization~\cite{chung2024depthreg3dgs} and decoupled appearance modules~\cite{lin2024vastgaussian}.
Recent works also investigate alternatives for updating the Gaussian parameters during training by extending the Adam optimizer~\cite{kingma2014adam} to account for visibility~\cite{taming3dgs} or by replacing it with second-order optimization algorithms~\cite{hoellein20253dgslm, lan2025secondordergs}.
Furthermore, adaptive Gaussian scheduling~\cite{chen2025dashgaussian} improves optimization speeds and feed-forward methods rely on large, pretrained models to fully avoid per-scene optimization and reduce reliance on dense input images~\cite{xu2025depthsplat, jiang2025anysplat, charatan2024pixelsplat, szymanowicz2024splatterimage, chen2025mvsplat, tang2024lgm}.

\paragraph{Representation and Applications}
Efficiency and portability, especially to allow for the reconstruction of larger scenes also recently gained interest.
Approaches in this area reason about the importance of Gaussians and their attributes to adjust how they are stored~\cite{3DGSzip2024} and loaded~\cite{zoomers2025progs}, or apply level-of-detail techniques to enable fast and high-quality rendering of large scenes~\cite{kerbl2024hierarchical, yang2025v3dg, kulhanek2025lodge, windisch2025lodofgaussians, ren2025octreegs}.
Significant effort has also been invested in extending 3DGS to non-rigid, \ie, dynamic scenes~\cite{yang2024deformable3dgs, wu20244dgs, yang20244dgaussians, luiten2023dynamic, xu2024temporalgaussianhierarchy, wang2025freetimegs, huang2025echoesofcoliseum}, efficient editing of trained models~\cite{schuetz2025splatshop, chen2024gaussianeditor, luo2025gaussianvrediting}, and meshing~\cite{guedon2024sugar, huang20242d, yu2024gof, radl2025sof, chen2025pgsr, zhang2024radegs, guedon2025milo}.
Beyond the original 3D Gaussian-based representation, there are also multiple works that build on the underlying pipeline proposed by Kerbl~\etal~\cite{kerbl3Dgaussians}, but use different primitives to \eg improve surface reconstruction~\cite{huang20242d, ye2025gaussiansurfel}, the representation of sharp edges~\cite{held2025convexsplatting, held2025trianglesplatting, von2025linprim, liu2025betasplatting}, or enable exact, \ie, overlap-aware volumetric rendering~\cite{mai2024ever}.

\paragraph{GPU Optimization}
In this work, we exploit various GPU optimization techniques to accelerate training, which allows us to achieve significant speedups over previously introduced techniques.
GPUs follow a SIMT (single-instruction multiple-threads) compute paradigm with individual small kernels, which is computationally fast due to the high amount of parallelism in the system~\cite{hijma2023optimizing}.
Kernel efficiency is commonly determined by the highest throughput. 
Typically, kernels are either memory-bound (spending most of their time waiting for or fetching data) or compute-bound, where the arithmetic instructions take the majority of the time.
Based on this, different optimization techniques can be used, such as exploiting GPU shared memory to allow multiple threads to load and access data efficiently and share costs.
For details, see the comprehensive survey of Hijma~\etal~\cite{hijma2023optimizing}.

%% file: sections/3_method.tex
\section{Method}
\label{sec:method}
We introduce a high-performance 3DGS optimization framework, \textit{Faster-GS}, which follows the same paradigm as the original method~\cite{kerbl3Dgaussians} with significant speed increases.
We first describe the scope and basis for this work (\cref{ssec:perf_review}), then consolidate and review recent optimization techniques for 3DGS (\cref{ssec:improvements}), and integrate further improvements (\cref{ssec:own_optim}).
Lastly, we present easy integration into 4D Gaussian Splatting (\cref{ssec:dynamic_extension}).

\subsection{Scope and Basis}
\label{ssec:perf_review}
\paragraph{Gaussian Splatting Performance}
Rendering splats with large screen space extensions involves scattered memory writes and is a common performance issue compared to, \eg, pixel-sized splats~\cite{schutz2022software,ruckert2022adop,franke2024trips}.
Kerbl~\etal~\cite{kerbl3Dgaussians} circumvent this problem by using a tile-based software rasterizer, which splits the image-plane into 16\texttimes16 tiles and intersects the bounding box of each splat with them, duplicating Gaussians for later stages into per-tile splat lists while streamlining memory.
Processing the splat lists is severely memory-bound, especially due to the high number of floats necessary for each Gaussian (see \cref{ssec:3dgs_recap}), which need to be loaded.
As such, reducing memory and memory accesses is the predominant way to accelerate 3DGS optimization, which we will discuss in the following sections.

\paragraph{Scope of this Work}
Our objective is to adhere closely to the original 3DGS optimization and outcomes, facilitating easy integration into existing works as well as prevalent frameworks~\cite{kerbl3Dgaussians,ye2025gsplat,lichtfeld2025}. 
We avoid extensive pruning or culling during reconstruction, which can enhance performance; however, it incurs (minimal) quality losses~\cite{radl2024stopthepop, hanson2025speedysplat, fang2025minisplatting, girish2024eagles}.
Additionally, we avoid compression, as it requires a careful trade-off between quality and compute~\cite{3DGSzip2024}.
However, while outside the scope of this work, further integration of pruning or compression should greatly increase speeds further~\cite{hanson2025speedysplat}.

\paragraph{Basis Implementation}
\label{ssec:basis_impl}
For a clean testbed and open-source version, we developed a refactored 3DGS implementation with several improvements aimed at enhancing numerical stability, efficiency, and modularity.
In particular, using front-to-back alpha blending for the backward pass (\cref{eq:alpha_blending}) removes the need for division-by-zero checks, and refined handling of degenerate quaternions of Gaussian covariances stabilizes gradients. 
Furthermore, explicit handling of $\mu_\text{2D}$ gradients and visibility masks reduces VRAM overhead.
For further details, see \cref{ssec:testbed_basis}. 
This \textit{basis} version increases performance by about 15\% compared to Kerbl~\etal~\cite{kerbl3Dgaussians}.

\subsection{A Survey of Recent Improvements}
\label{ssec:improvements}

In the following, we compile and group techniques from recent 3DGS literature to address memory bottlenecks, ensuring no negative impact on reconstruction quality. We highlight required contributions to the training pipeline and integrate minor novel optimizations in them.

\paragraph{Splat Bounding}
\label{par:splat_bounding}
The splatting algorithm of \cref{eq:mean_projection,eq:cov_projection} results in a 2D Gaussian on the image plane.
3DGS skips all fragments created during rasterization where $\alpha$ (see \cref{eq:alpha_computation}) is below $\tau_\alpha\!=\!\nicefrac{1}{255}$.
This corresponds to truncating the Gaussian at roughly $3.33\sigma$ and allows bounding the relevant area of the splat with a 2D ellipse.
As efficient tile-based rendering requires creating per-tile lists of all contributing splats, the bounding box of this 2D ellipse is of interest.
In 3DGS, Kerbl~\etal use a square-shaped bounding box, which -- as previously discussed by Radl~\etal~\cite{radl2024stopthepop} -- underestimates the size of this bounding box for opaque, axis-aligned splats.
This is because each splat is bound with an axis-aligned square around the circle with radius $3\sigma$ instead of $3.33\sigma$, the value corresponding to $\tau_\alpha$.
A natural improvement used in prior works~\cite{radl2024stopthepop, wang2024adrgaussian, hanson2025speedysplat} is to bound the splats with an axis-aligned rectangle instead of a square, where the center of the rectangle is at $\mu_\text{2D}$ and its half-extents are given by $\sqrt{\Sigma_{\text{2D}_{1,1}}}$ and $\sqrt{\Sigma_{\text{2D}_{2,2}}}$ respectively.

The opacity $o$ of the Gaussian can also be factored in by multiplying $-2\ln(\nicefrac{\tau_\alpha}{o})$ with the radicand before applying the square root.
This factor follows from setting \cref{eq:alpha_computation} equal to $\tau_\alpha$ and solving for the numerator of the exponent.
This full, opacity-aware formulation leads to a notable decrease in false positives included in each per-tile splat list.

\paragraph{Tile-based Culling}
While the aforementioned tight rectangles are the optimal axis-aligned bounding box, they can still overestimate tile intersections for some splats.
To eliminate this remaining overhead, Radl~\etal~\cite{radl2024stopthepop} and Hanson~\etal~\cite{hanson2025speedysplat} propose algorithms for efficiently computing what tiles each ellipse overlaps with.
Hanson~\etal iterate over the shorter side of the bounding rectangle and determine the first and last tile a splat may overlap with in each row$\slash$column, which minimizes the number of computations but may cause warp divergence when the rectangle sizes are very different.
In contrast, Radl~\etal implement a load-balanced approach for checking all tiles inside the bounding rectangle by computing the point where the value of the Gaussian is maximal for each tile.
For our implementation, we select the approach by Radl~\etal as we determined it to be faster due to the simpler control flow and added load balancing, but note that the two approaches are not mutually exclusive.

\paragraph{Sorting}
\label{par:separate_sorting}
To create the per-tile splat lists used for the tiled rasterization approach, 3DGS writes key$\slash$value pairs for each tile$\slash$Gaussian pair.
The original implementation uses 64-bit keys, where the most significant 32 bits indicate the tile index and the 32 least significant ones contain the bits of $\hat{\mu}_z$, \ie, depth information.
After writing these key$\slash$value pairs to a large buffer, they are sorted using a single radix sort to obtain depth-sorted lists of splats for each tile.
Recent work by Schütz~\etal~\cite{schuetz2025splatshop} shows that separating this sorting step into two stages, one to establish depth ordering and a second to obtain per-tile lists, reduces VRAM usage as well as the total time spent on sorting.
Note that this change requires using a stable sorting algorithm, \eg, radix sort.
See \cref{ssec:separate_sorting_details} for details.

\paragraph{Per-Gaussian Backward}
\label{par:per_gaussian_backward}
In the original 3DGS implementation, by far the most expensive operation in terms of runtime is the backward pass computing the alpha blending gradients.
This is because each splat may contribute to an arbitrary number of pixels, which introduces the need for using atomic operations for accumulation.
As explored by Durvasula~\etal~\cite{durvasula2023distwar}, this is computationally suboptimal, which they solve through custom atomic functions.

Recent work by Mallick~\etal~\cite{taming3dgs} avoids this problem by parallelizing over Gaussians instead of pixels in the backward pass, which reduces the number of required atomic operations by a factor equal to the number of pixels in each tile, \ie, 256 as usually a tile size of 16\texttimes16 pixels is used for 3DGS.
For efficiency, they store the alpha blending state at each non-empty pixel after every 32\textsuperscript{nd} splat in the respective tile list during the forward pass.
While this approach speeds up training significantly by addressing a major bottleneck, it is the only addition that increases VRAM usage.

We integrate and improve the design of Mallick~\etal~\cite{taming3dgs} by exploiting shared memory to further reduce memory costs and reduce overall VRAM allocations (see \cref{ssec:per_gaussian_backward_details}).

\paragraph{Rasterization Kernel Fusion}
While the PyTorch-based frontend of the original 3DGS implementation makes it very flexible and easy to extend, most instructions are set up as non-fused, individual CUDA kernels, which frequently have to load and store buffers.
Two methods for mitigating this are employed:
First, prior works~\cite{taming3dgs, hahlbohm2025htgs} skip the concatenation of the two SH coefficient buffers (3DGS stores these separately to allow for different learning rates of the view-independent and view-dependent bands) and pass these buffers separately to the rasterization backend.
This fuses the concatenation into the rasterization kernel and has positive effects on performance (see \cref{sec:evaluation}).
Second, we fuse the activation functions for the scales, rotations, and opacities of the Gaussians into the rasterizer to avoid any PyTorch-related overhead.
This is already commonly used to speed up inference rendering~\cite{hahlbohm2025htgs,schuetz2025splatshop} and for training we can also fuse the gradient computations required for added benefits.

\subsection{Refining Optimized Implementations}
\label{ssec:own_optim}
Integrating the previously surveyed optimizations results in a strong reduction in memory costs and improved performance.
We further investigate optimization techniques, adapting this optimized baseline.

\paragraph{Parameter Updates}
As we will show in our experiments, a surprisingly expensive part of the training pipeline is the optimizer steps, which adjust each Gaussian's parameters.
We find that the main performance issue with the original 3DGS implementation \wrt to the optimizer originates from the use of a non-fused Adam update routine.
Recent versions of PyTorch allow users to avoid this by passing \verb|fused=True|.
A slightly faster alternative involves using FusedAdam from the NVIDIA apex library as a drop-in replacement.

However, for further increased performance, we develop our own Adam implementation in which we precisely match the behavior of the PyTorch implementation while optimizing away all unnecessary overhead.
It fully exploits the fusion of the kernel into CUDA, together with fast math operations and fewer overall instructions through fused-multiply additions, further accelerating the optimization.

\paragraph{Locality-preserving Densification}
Through the strong reduction of memory costs, we find that memory layout becomes a throttling factor.
During densification, new Gaussians are added at the end of the parameter buffers, which causes spatially close Gaussians to be far apart in memory, which results in uncoalesced memory accesses, as neighboring threads need to access different parts of the memory.

To better align Gaussians, we introduce a simple addition to
training when densification is active.
We regularly apply \textit{z}-ordering~\cite{SCHUETZ-2021-PCC} to the current set of Gaussians to ensure neighboring Gaussians in 3D are also close inside the parameter buffers.
This reduces warp divergence and cache misses, resulting in faster training when scenes contain many Gaussians.
While \textit{z}-ordering is computationally efficient (roughly 4 ms per million Gaussians), we find that frequent application has diminishing returns. We empirically determined performing this step every 5000 iterations works well across scenes.
We further observe that it performs effectively only when used with the per-Gaussian backward pass, as numerous atomic operations in the original backward pass would otherwise heighten atomic contention.

\paragraph{Fusing Backward and Optimizer}
\label{par:fused_bw_opt}
As we will show in our experiments, applying all aforementioned improvements leads to a major speedup over the original 3DGS implementation.
When profiling the performance of the resulting framework, we find that the GPU spends between 40\% and 60\% of the total training time on the optimized Adam update routine.
To alleviate this bottleneck, we fuse the parameter updates directly into the backward pass of the rasterization module by first loading moments and computing all parameter updates during gradient computation.
This reduces VRAM requirements (especially for large amounts of Gaussians) as no additional buffers for the parameters are necessary.
However, to maintain correctness \wrt the Adam step, we need to perform parameter updates for parameters receiving a gradient of zero in a given iteration, \eg, due to their Gaussian being outside the viewing frustum.
This reduces attainable performance improvements with this fused approach.

A recent idea by Mallick~\etal~\cite{taming3dgs} is to skip updates for these invisible Gaussians, which fits our fused design exceptionally well.
We see this as an optional extension for further acceleration, however, as it can cause inconsistencies and performance regressions compared to the original 3DGS implementation (see our evaluation and Mallick~\etal~\cite{taming3dgs}).

\subsection{Extension to Dynamic Scenes}
\label{ssec:dynamic_extension}
Managing dynamic objects in a scene is a critical issue in 3D reconstruction.
Although these elements can occasionally be considered distractors~\cite{sabour2025spotlesssplats}, the dynamic object, such as a human, frequently constitutes the most relevant aspect of a scene.
We integrate our performance-optimized Gaussian Splatting framework to support optimization of 4D Gaussians based on the approach by Yang~\etal~\cite{yang20244dgaussians}.
A 4D Gaussian is constructed analogously to a 3D Gaussian (\cref{ssec:3dgs_recap}), with the addition of two parameters accounting for the mean and scale along the temporal dimension. The 4D rotation is separated into a left-isoclinic and a right-isoclinic rotation, each represented by a quaternion.
For rendering a given timestep $t$, Yang~\etal compute the conditional 3D Gaussian:
\begin{align}
\mu_{\text{3D}|t} &= \mu_{1:3} + \Sigma_{1:3,4}\Sigma_{4,4}^{-1}(t-\mu_4),\\
\Sigma_{\text{3D}|t} &= \Sigma_{1:3,1:3} - \Sigma_{1:3,4}\Sigma_{4,4}^{-1}\Sigma_{4,1:3}.
\end{align}
In combination with the value of the marginal distribution, \ie, the 1D Gaussian $p(t)\!=\!\mathcal{N}(t;\mu_4, \Sigma_{4,4})$ evaluated at $t$, multiplied by the result of \cref{eq:alpha_computation}, Yang~\etal develop a differentiable approach for 4D Gaussian rendering.

We integrate this approach into our optimized 3DGS framework by adapting the data model accordingly and extending the rasterizer kernels to compute the conditional$\slash$marginal Gaussians and the relevant gradients in the forward and backward pass respectively.
We also extend our training schedule to match that of Yang~\etal, who render and propagate the loss for multiple images in each training iteration.
We note that with this, the previous optimizations are directly transferable to 4D Gaussian optimization.

%% file: sections/4_evaluation.tex
\section{Evaluation}
\label{sec:evaluation}

We evaluate our developed framework in a comprehensive suite of experiments with three main goals.
Confirming that the quality has not regressed relative to pertinent baselines, examining the speed enhancements related to each addition both individually and collectively, and analyzing the extension of our non-rigid reconstruction method.

\subsection{Setup}
\label{ssec:experiment_setup}
As baselines, we compare with the official 3DGS implementation by Kerbl~\etal~\cite{kerbl3Dgaussians}, the 3DGS implementation of Radl~\etal with tight, opacity-aware bounding boxes and load-balanced tile-based culling~\cite{radl2024stopthepop}, a variant of Speedy-Splat~\cite{hanson2025speedysplat} only using the SnugBox and AccuTile features, and the 3DGS-accel branch of the official 3DGS codebase.
The latter is effectively identical to the 3DGS implementation from Taming-3DGS~\cite{taming3dgs}, which uses opacity-aware tile-based culling without load balancing, per-Gaussian backwards, and separate SH buffers within the rasterizer.
For all methods, we use the fused SSIM implementation by Goel~\etal as proposed in Taming-3DGS~\cite{taming3dgs} during loss computation.
We unify hyperparameters across all methods, which is necessary following a recent change in the official 3DGS codebase.
Training images are uploaded to VRAM before optimization, which is excluded from the reported training times but included in peak VRAM.
Unless otherwise noted, all experiments were conducted on the same hardware using a single RTX 4090 GPU.
We use the standard benchmark for 3DGS methods, \ie, 13 scenes from the Mip-NeRF360~\cite{barron2022mipnerf360}, Tanks and Temples~\cite{Knapitsch2017}, and Deep Blending~\cite{hedman2018deep} datasets with a 7:1 train$\slash$test split.
Image quality metrics (PSNR, SSIM, and LPIPS~\cite{zhang2018lpips}) are computed under identical conditions, \ie, with the same script, where we ensure a correct LPIPS computation by normalizing images to $[-1, 1]$.
We also use pre-downscaled images for training and testing when these are provided with the dataset~\cite{barron2022mipnerf360}.

\input{tables/1_quantitative_comparisons}

\subsection{Results}
In \cref{tab:main_results}, we show averaged results for baselines and our implementation.
As expected, all methods achieve the same image quality and optimize to roughly the same number of Gaussians.
Note that image quality results can vary significantly between runs, even when using the same fixed random seed due to the interaction between floating point math and random ordering during gradient accumulation.
For some scenes, \eg, Bonsai from Mip-NeRF360~\cite{barron2022mipnerf360}, this can lead to a major difference of up to 0.5 dB PSNR in all implementations.
Thus presented image quality metrics are averaged across five runs.
The main difference between methods is in the training time and VRAM consumption, where our implementation outperforms all baselines significantly, by up to 5.2\texttimes\ compared to 3DGS~\cite{kerbl3Dgaussians} and 2.4\texttimes\ compared to Taming-3DGS~\cite{taming3dgs} on the Deep Blending scenes.
Notably, our basis implementation also achieves strong results.
\input{figures/1_runtime_comparisons}

\paragraph{Individual Components}
\input{tables/2_ablations}
\input{tables/2_optimizer_bottleneck}
Starting from our basis implementation, we show the isolated performance impact of all changes in \cref{tab:ablations}.
We report results for the indoor and outdoor scenes of the Mip-NeRF360~\cite{barron2022mipnerf360} separately due to different optimization behavior.
Specifically, the number of Gaussians created during densification are a lot higher for outdoor scenes (3.8M vs.\ 1.3M Gaussians on average), while indoor scenes use images with roughly 1.5\texttimes\ more pixels.

Particularly impressive speedups are obtained by using any of the three fused Adam techniques, with our implementation consistently outperforming those from PyTorch and apex.
The per-Gaussian backward pass for alpha blending also leads to a major speedup -- especially when the number of primitives is small -- but at the same time is the only change that increases VRAM usage.

We also find that the load-balanced approach for creating Gaussian$\slash$tile instances has an increasingly negative impact on training speed as the number of primitives increases.
Note that full tile-based culling is also affected.
When analyzing this, we found the underlying issue to be warp divergence, where more than half of the threads in each warp are inactive because their Gaussian is invisible from the current viewpoint.
While load balancing is meant to address precisely this issue, we find that the associated uncoalesced memory accesses are a bigger bottleneck.

We can resolve this issue by sorting Gaussians in \textit{z}-order at regular intervals during training.
While this works great for scenes with many Gaussians, it significantly slows down training when the number of Gaussians is small due to a massive increase in atomic contention in the alpha blending backward pass.
Specifically, threads in different tiles are more likely to write to the same cache line, which reduces performance as operations are serialized and cache lines are invalidated after every write (a.k.a.\ false sharing).
Fortunately, this is much less of an issue with the per-Gaussian backward pass enabled in our full implementation, where omitting our repeated \textit{z}-ordering during densification slows down training across all scene subsets.

In \cref{fig:runtime_comparisons}, we investigate the time spent per algorithm step and find that computing parameter updates is a significant fraction of the runtime.
We evaluate approaches for decreasing this bottleneck in \cref{tab:optimizer_bottleneck}.
Fusing the optimizer with the backward pass of the rasterizer (see \cref{par:fused_bw_opt}) provides a small speedup and VRAM reduction over our full version but reduces its extensibility.
After integration, parameter updates remain the main bottleneck of training.
Skipping all parameter updates for invisible Gaussians~\cite{taming3dgs} or omitting view-dependent SH coefficients speeds up training and reduces VRAM usage, but it also slightly degrades quality.

\input{tables/3_gpu_comparison}
\paragraph{GPU Comparison}
We investigate the advantages of our improved 3DGS framework across different GPUs from the three most recent consumer-grade generations of Nvidia GPUs.
As shown in \cref{tab:gpu_impact}, newer GPUs exhibit greater speedup, suggesting anticipated performance improvements in upcoming hardware generations.
On an RTX 5090 GPU, training takes 163 seconds on average, a 5\texttimes\ improvement over the original implementation.

\subsection{Dynamic Scenes}
\input{tables/4_dnerf_comparison}
To evaluate our extension to 4D reconstruction (see \cref{ssec:dynamic_extension}), we compare our implementation against the reference implementation by Yang~\etal~\cite{yang20244dgaussians} on the eight synthetic scenes from the D-NeRF~\cite{pumarola2021dnerf} dataset.
All scenes are trained and evaluated at the native dataset resolution of 800\texttimes800 pixels using the provided train$\slash$test splits.
Note that we use one consistent set of hyperparameters across all scenes:
We initialize with 100K random points, use the standard view-dependent color parametrization from 3DGS (SH up to degree three), and train for 30000 iterations with a batch size of four.
The results in \cref{tab:dnerf_comparison} show that our speedup for standard 3DGS translates to the dynamic scene reconstruction setting as our improved implementation trains up to 3\texttimes\ faster while using less VRAM and maintaining quality.
See \cref{ssec:n3dv_results} for additional quantitative comparisons on real-world data.

%% file: tables/1_quantitative_comparisons.tex
\begin{table*}[tb]
\caption{%
Quantitative comparisons on the Mip-NeRF360, Tanks and Temples, and Deep Blending datasets. For baselines marked with $\dagger$ we enable only those contributions that do not affect quality. The three best results are highlighted in \textcolor{1stText}{\textbf{green}} in descending order of saturation.
}\label{tab:main_results}\vspace{-1mm}
\centering
\setlength\tabcolsep{1.2pt}
\scriptsize
\begin{tabular}{lcccccc|cccccc|cccccc}
\toprule
 & \multicolumn{6}{c|}{Mip-NeRF360~\cite{barron2022mipnerf360}} & \multicolumn{6}{c|}{Tanks and Temples~\cite{Knapitsch2017}} & \multicolumn{6}{c}{Deep Blending~\cite{hedman2018deep}} \\
Method & PSNR$^\uparrow$ & SSIM$^\uparrow$ & LPIPS$^\downarrow$ & Train$^\downarrow$ & VRAM$^\downarrow$ & \#Gs$^\downarrow$ & PSNR$^\uparrow$ & SSIM$^\uparrow$ & LPIPS$^\downarrow$ & Train$^\downarrow$ & VRAM$^\downarrow$ & \#Gs$^\downarrow$ & PSNR$^\uparrow$ & SSIM$^\uparrow$ & LPIPS$^\downarrow$ & Train$^\downarrow$ & VRAM$^\downarrow$ & \#Gs$^\downarrow$ \\ 
\midrule
3DGS \cite{kerbl3Dgaussians}                       & 27.53 & 0.815 & 0.256 & 18m44s & 8.8GiB & 2.74M & 23.77 & 0.852 & 0.204 & 11m26s & 4.7GiB & 1.57M & 29.81 & 0.907 & 0.305 & 19m43s & 8.1GiB & 2.47M \\
Speedy-Splat$^\dagger$\cite{hanson2025speedysplat} & 27.53 & 0.816 & 0.255 & 17m32s & \cellcolor{3rd}7.5GiB & 2.72M & 23.77 & 0.852 & 0.205 & 10m34s & \cellcolor{3rd}4.1GiB & 1.57M & 29.79 & 0.906 & 0.304 & 18m40s & \cellcolor{3rd}7.1GiB & 2.55M \\
StopThePop$^\dagger$\cite{radl2024stopthepop}      & 27.54 & 0.816 & 0.255 & 16m46s & 7.6GiB & 2.73M & 23.76 & 0.852 & 0.205 & 9m51s  & \cellcolor{3rd}4.1GiB & 1.57M & 29.83 & 0.907 & 0.304 & 17m47s & \cellcolor{3rd}7.1GiB & 2.55M \\
Taming-3DGS$^\dagger$\cite{taming3dgs}             & 27.53 & 0.815 & 0.256 & \cellcolor{2nd}10m49s & 8.9GiB & 2.73M & 23.78 & 0.852 & 0.203 & \cellcolor{2nd}7m04s  & 4.9GiB & 1.57M & 29.81 & 0.906 & 0.305 & \cellcolor{2nd}9m01s  & 8.4GiB & 2.47M \\
Basis Impl. [Ours]                                       & 27.57 & 0.816 & 0.255 & \cellcolor{3rd}15m57s & \cellcolor{2nd}6.3GiB & 2.67M & 23.79 & 0.853 & 0.204 & \cellcolor{3rd}9m39s  & \cellcolor{1st}3.4GiB & 1.52M & 29.74 & 0.907 & 0.304 & \cellcolor{3rd}17m15s & \cellcolor{1st}6.0GiB & 2.52M \\
Ours                                               & 27.56 & 0.816 & 0.254 & \cellcolor{1st}4m31s  & \cellcolor{1st}6.1GiB & 2.73M & 23.75 & 0.853 & 0.204 & \cellcolor{1st}3m04s  & \cellcolor{1st}3.4GiB & 1.55M & 29.78 & 0.906 & 0.304 & \cellcolor{1st}3m46s  & \cellcolor{1st}6.0GiB & 2.61M \\
\bottomrule
\end{tabular}
\end{table*}

%% file: figures/1_runtime_comparisons.tex
\begin{figure}[bt]
\centering
\includegraphics[width=\linewidth]{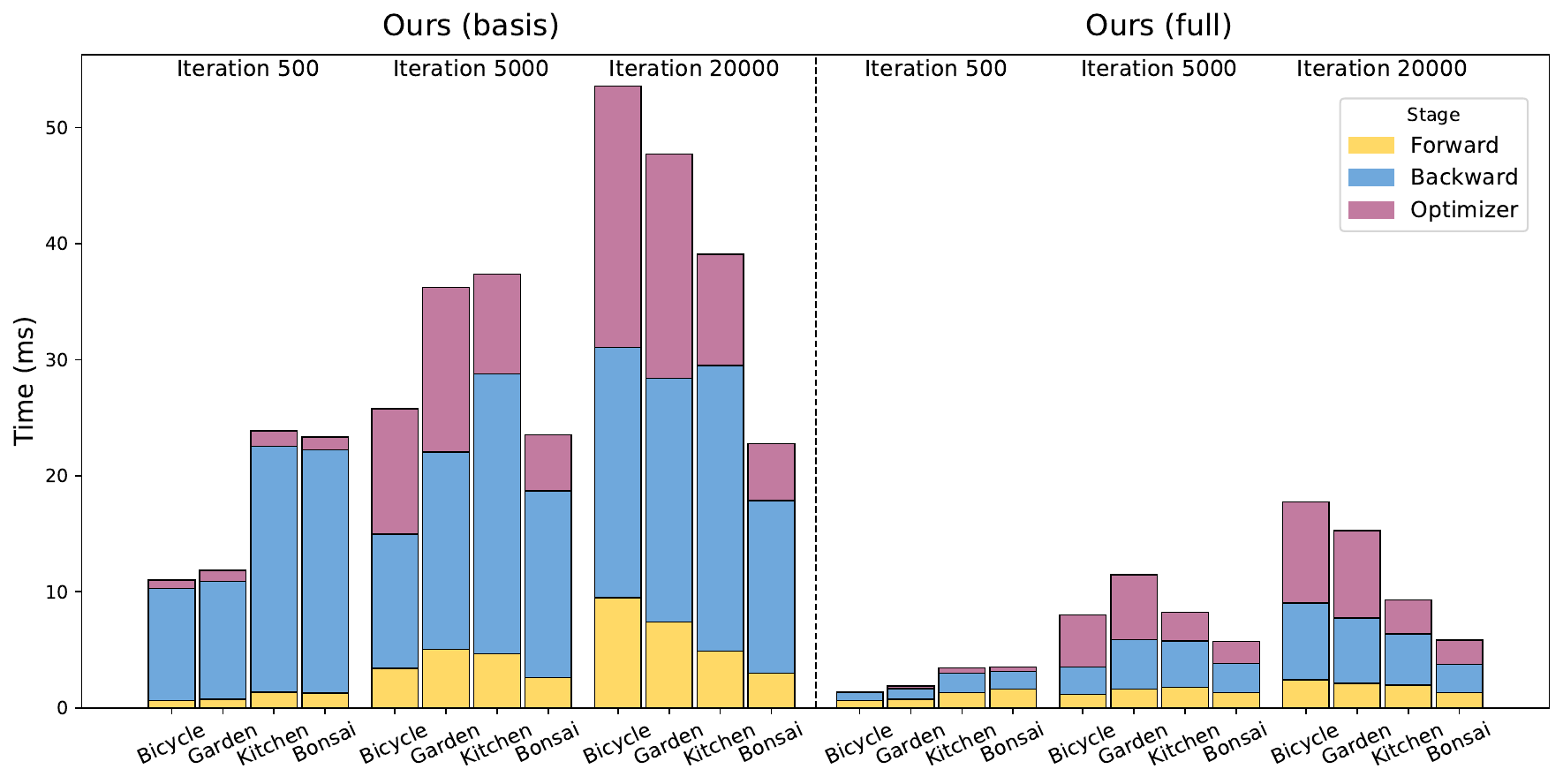}
\caption{%
Runtime comparison for the basis and full versions of our optimized 3DGS framework. We measure the time it takes to compute the forward$\slash$backward pass and the optimizer step respectively during iteration 500, 5000, and 20000 when training four scenes from the Mip-NeRF360 dataset~\cite{barron2022mipnerf360}.
}\label{fig:runtime_comparisons}
\end{figure}

%% file: tables/2_ablations.tex
\begin{table*}[tb]
\caption{%
Ablations on the Mip-NeRF360, Tanks and Temples, and Deep Blending datasets using an RTX 4090 GPU. Relative improvements were computed before rounding and indicate the speedup$\slash$reduction in training time and peak VRAM usage respectively.
}\label{tab:ablations}\vspace{-1mm}
\centering
\setlength\tabcolsep{2.9pt}
\scriptsize
\begin{tabular}{lcc|cc|cc|cc}
\toprule
 & \multicolumn{2}{c|}{Mip-NeRF360~\cite{barron2022mipnerf360} - Outdoor} & \multicolumn{2}{c|}{Mip-NeRF360~\cite{barron2022mipnerf360} - Indoor} & \multicolumn{2}{c|}{Tanks and Temples~\cite{Knapitsch2017}} & \multicolumn{2}{c}{Deep Blending~\cite{hedman2018deep}} \\
Method & Training$^\downarrow$ & VRAM$^\downarrow$ & Training$^\downarrow$ & VRAM$^\downarrow$ & Training$^\downarrow$ & VRAM$^\downarrow$ & Training$^\downarrow$ & VRAM$^\downarrow$ \\ 
\midrule
Basis                                   & 17m07s & 6.39GiB & 14m29s & 6.23GiB & 9m39s  & 3.43GiB & 17m15s & 6.04GiB \\
~~+ fused activations                   & 16m19s~(1.05\texttimes) & 6.27GiB~(0.98\texttimes) & 14m00s~(1.03\texttimes) & 6.19GiB~(0.99\texttimes) & 9m19s~(1.04\texttimes) & 3.37GiB~(0.98\texttimes) & 16m25s~(1.05\texttimes) & 5.94GiB~(0.98\texttimes) \\
~~+ separate SH buffers                 & 15m39s~(1.09\texttimes) & 5.72GiB~(0.89\texttimes) & 14m10s~(1.02\texttimes) & 5.99GiB~(0.96\texttimes) & 9m13s~(1.05\texttimes) & 3.13GiB~(0.91\texttimes) & 16m31s~(1.04\texttimes) & 5.57GiB~(0.92\texttimes) \\
~~+ rectangular AABBs                   & 16m52s~(1.02\texttimes) & 6.39GiB~(1.00\texttimes) & 13m58s~(1.04\texttimes) & 6.16GiB~(0.99\texttimes) & 9m17s~(1.04\texttimes) & 3.38GiB~(0.99\texttimes) & 16m44s~(1.03\texttimes) & 5.97GiB~(0.99\texttimes) \\
~~+ rect.\ AABBs w$\slash$ opacity & 16m46s~(1.02\texttimes) & 6.38GiB~(1.00\texttimes) & 13m47s~(1.05\texttimes) & 6.12GiB~(0.98\texttimes) & 9m13s~(1.05\texttimes) & 3.36GiB~(0.98\texttimes) & 16m44s~(1.03\texttimes) & 5.92GiB~(0.98\texttimes) \\
~~+ load-balanced instancing     & 17m10s~(1.00\texttimes) & 6.40GiB~(1.00\texttimes) & 14m22s~(1.01\texttimes) & 6.23GiB~(1.00\texttimes) & 9m35s~(1.01\texttimes) & 3.42GiB~(1.00\texttimes) & 17m07s~(1.01\texttimes) & 6.04GiB~(1.00\texttimes) \\
~~+ full tile-based culling             & 16m53s~(1.01\texttimes) & 6.39GiB~(1.00\texttimes) & 13m40s~(1.06\texttimes) & 6.11GiB~(0.98\texttimes) & 9m11s~(1.05\texttimes) & 3.35GiB~(0.98\texttimes) & 16m43s~(1.03\texttimes) & 5.89GiB~(0.97\texttimes) \\
~~+ separate sorting                    & 16m56s~(1.01\texttimes) & 6.33GiB~(0.99\texttimes) & 14m05s~(1.03\texttimes) & 6.12GiB~(0.98\texttimes) & 9m34s~(1.01\texttimes) & 3.36GiB~(0.98\texttimes) & 16m58s~(1.02\texttimes) & 5.90GiB~(0.98\texttimes) \\
~~+ per-Gaussian backward               & 14m14s~(1.20\texttimes) & 7.69GiB~(1.20\texttimes) & ~~7m52s~(1.84\texttimes)& 7.84GiB~(1.26\texttimes) & 6m56s~(1.39\texttimes) & 4.49GiB~(1.31\texttimes) & 10m27s~(1.65\texttimes) & 8.36GiB~(1.38\texttimes) \\
~~+ fused Adam (PyTorch)                & 14m03s~(1.22\texttimes) & 6.42GiB~(1.00\texttimes) & 13m40s~(1.06\texttimes) & 6.23GiB~(1.00\texttimes) & 8m40s~(1.11\texttimes) & 3.42GiB~(1.00\texttimes) & 15m22s~(1.12\texttimes) & 6.05GiB~(1.00\texttimes) \\
~~+ fused Adam (Apex)                   & 13m12s~(1.30\texttimes) & 6.41GiB~(1.00\texttimes) & 12m59s~(1.12\texttimes) & 6.22GiB~(1.00\texttimes) & 8m02s~(1.20\texttimes) & 3.42GiB~(1.00\texttimes) & 14m38s~(1.18\texttimes) & 6.04GiB~(1.00\texttimes) \\
~~+ fused Adam (Ours)                   & 12m50s~(1.33\texttimes) & 6.40GiB~(1.00\texttimes) & 12m55s~(1.12\texttimes) & 6.23GiB~(1.00\texttimes) & 7m55s~(1.22\texttimes) & 3.42GiB~(1.00\texttimes) & 14m32s~(1.19\texttimes) & 6.04GiB~(1.00\texttimes) \\
Full w$\slash$o \textit{z}-ordering              & ~~5m52s~(2.92\texttimes)& 5.99GiB~(0.94\texttimes) & ~~3m17s~(4.41\texttimes)& 6.25GiB~(1.00\texttimes) & 3m11s~(3.02\texttimes) & 3.36GiB~(0.98\texttimes) & ~~3m50s~(4.50\texttimes)& 5.95GiB~(0.98\texttimes) \bstrut\\
\hline
Full                                    & ~~5m31s~(3.10\texttimes)& 5.99GiB~(0.94\texttimes) & ~~3m14s~(4.47\texttimes)& 6.29GiB~(1.01\texttimes) & 3m04s~(3.14\texttimes) & 3.39GiB~(0.99\texttimes) & ~~3m46s~(4.58\texttimes)& 5.96GiB~(0.99\texttimes) \tstrut\\
\bottomrule
\end{tabular}
\end{table*}

%% file: tables/2_optimizer_bottleneck.tex
\begin{table}[tb]
\caption{%
Integrating the optimizer step into the backward pass results in a minor speed enhancement. Avoiding updates for non-visible Gaussians or excluding view-dependent spherical harmonics (SH) affects quality. Results are averaged over the five outdoor scenes from the Mip-NeRF360 dataset~\cite{barron2022mipnerf360}.
}\label{tab:optimizer_bottleneck}%
\centering
\setlength\tabcolsep{3.6pt}
\footnotesize
\begin{tabular}{lcccc}
\toprule
 & PSNR$^\uparrow$ & Train$^\downarrow$ & VRAM$^\downarrow$ & \#Gs$^\downarrow$ \\
\midrule
Full                                & 24.72 & 5m31s & 6.0GiB & 3.87M \\
~~+ fused updates                   & 24.73 & 5m04s & 5.6GiB & 3.89M \\
~~+ fused updates (skip invisible)  & 24.59 & 3m03s & 5.1GiB & 3.34M \\
~~+ fused updates (SH degree${=}$0) & 24.38 & 2m24s & 3.2GiB & 3.85M \\
\bottomrule
\end{tabular}
\end{table}

%% file: tables/3_gpu_comparison.tex
\begin{table}[tb]
\caption{%
Training time across all 13 scenes with different GPUs.
}\label{tab:gpu_impact}%
\centering
\setlength\tabcolsep{9pt}
\footnotesize
\begin{tabular}{lccc}
\toprule
                             & RTX 3090              & RTX 4090              & RTX 5090              \\
\midrule
3DGS \cite{kerbl3Dgaussians} & 23m46s                & 17m46s                & 13m05s                \\
Ours (Full)                         & 6m03s (3.9\texttimes) & 4m10s (4.3\texttimes) & 2m43s (4.8\texttimes) \\
\bottomrule
\end{tabular}
\end{table}

%% file: tables/4_dnerf_comparison.tex
\begin{table}[tb]
\caption{%
Comparison with the reference implementation~\cite{yang20244dgaussians} for our extension to 4D Gaussians on the synthetic D-NeRF dataset~\cite{pumarola2021dnerf}.
}\label{tab:dnerf_comparison}%
\centering
\setlength\tabcolsep{1.6pt} %
\footnotesize
\begin{tabular}{lcccccc}
\toprule
 & PSNR$^\uparrow$ & SSIM$^\uparrow$ & LPIPS$^\uparrow$ & Train$^\downarrow$ & VRAM$^\downarrow$ & \#Gs$^\downarrow$ \\
\midrule
Yang~\etal~\cite{yang20244dgaussians} & 31.52 & 0.960 & 0.051 & \cellcolor{2nd}18m09s & \cellcolor{2nd}1.9GiB & 0.83M \\
Ours                                  & 31.79 & 0.960 & 0.051 & \cellcolor{1st}6m22s (2.8\texttimes) & \cellcolor{1st}1.2GiB & 0.79M \\
\bottomrule
\end{tabular}
\end{table}

%% file: sections/5_discussion.tex
\section{Discussion, Limitations, and Future Work}
\label{sec:discussion}
Our optimized framework significantly accelerates Gaussian Splatting and we find that the remaining bottlenecks are tied to the computation of parameter updates.
This motivates the integration of second-order optimization algorithms~\cite{hoellein20253dgslm, lan2025secondordergs} or more compact view-dependent appearance representations~\cite{yariv2023bakedsdf}, which we leave as future work.
We also highlight that further optimizations, \eg, fusing the forward and backward passes~\cite{poirier2025editable} or mixed precision training~\cite{mueller2022instant}, are possible but will come with a tradeoff between simplicity, robustness, and optimal performance.
While not within the scope of our work, we present results for integrating state-of-the-art anti-aliasing and densification techniques~\cite{yu2024mip, kheradmand243dgsmcmc} as well as an inference-optimized rasterizer implementation based on our testbed to facilitate this process (see \cref{ssec:inference_fps,ssec:aa_results,ssec:mcmc_results}).
Furthermore, our evaluation sets aside valuable training improvements \wrt to artifacts~\cite{radl2024stopthepop}, controllability~\cite{taming3dgs}, and informed pruning techniques~\cite{hanson2025speedysplat} that could be added in the future.

%% file: sections/6_conclusion.tex
\section{Conclusion}
\label{sec:conclusion}
In this paper, we surveyed recent 3DGS follow-up works for performance improvements and systematically evaluated their effectiveness.
We further integrate memory-efficient adaptations to arrive at a new, optimized Gaussian Splatting framework, \textit{Faster-GS}, that trains 3D and 4D Gaussian scenes up to 5\texttimes\ faster than prior work.
Furthermore, we reduce VRAM requirements by up to 30\%, making our approach especially cost-effective and feasible to use on lower-end hardware.
Ultimately, our framework enables full 3DGS reconstruction in less than two minutes.
With its code release, we hope to significantly accelerate future Gaussian Splatting-based view synthesis research.

\section*{Acknowledgments}
This work was partially funded by the DFG projects ``Real-Action VR'' (ID 523421583) and ``Increasing Realism of Omnidirectional Videos in Virtual Reality'' (ID 491805996).
Linus Franke was supported by the ERC Advanced Grant NERPHYS (ID 101141721).%

%% file: sections/X_suppl.tex
\section{3DGS Training Details}
\label{sec:training_details}
To complement our brief summary of the 3DGS training schedule (\cref{ssec:3dgs_recap}), we supplement the remaining details and all relevant hyperparameter values in the following.
Note that the specified hyperparameters precisely reflect the optimization schedule of the original implementation by Kerbl~\etal~\cite{kerbl3Dgaussians}.
Recall that to reconstruct a scene from a set of training images, 3DGS first initializes Gaussians at random positions or based on an input point cloud.
As this initial point cloud is often very sparse, Kerbl~\etal~\cite{kerbl3Dgaussians} introduce adaptive density control (ADC), a set of carefully tuned heuristics for cloning, splitting, and pruning Gaussians during optimization.
Specifically, 3DGS tracks the magnitude of the $\mu_\text{2D}$ gradients during training for all Gaussians.
At regular intervals, Gaussians for with the average magnitude of this gradient exceeds a predefined threshold are selected for densification.
Of the selected Gaussians, large ones are each split into two new, smaller Gaussians with new positions being sampled from the respective parent Gaussian distributions, while those that are small are simply duplicated in place.
Additionally, Gaussians that either have a very low opacity or are too large \wrt scene size are removed.
Densification starts after a warm-up period of 600 iterations and is repeated every 100 iterations thereafter.
Gaussians are split$\slash$cloned when their average positional gradient across all iterations where they were visible (\ie, inside the viewing frustum) since the previous densification step exceeds $2\text{e}\!-\!4$, and pruned when their opacity is below 0.05.
After the final densification step in iteration 14900, Gaussians are no longer added or removed from the model.

To further encourage pruning of floaters and incorrectly placed Gaussians, 3DGS resets the opacity of all Gaussians to a small value multiple times during the optimization.
The opacity reset is performed every 3000 iterations while densification is active, \ie, four times in total and clips opacity values to 0.01 from above.

For the loss function, 3DGS uses a weighted combination of L1 and D-SSIM (\ie $1-\text{SSIM}$) terms, with weights being 0.8 and 0.2 respectively.
Parameter updates are then performed using the ADAM optimizer~\cite{kingma2014adam} with $\beta_1\!=\!0.9$, $\beta_2\!=\!0.999$, and $\epsilon\!=\!1\text{e}\!-\!15$.
Learning rates are set to 0.005, 0.001, and 0.025 for scale, rotation, and opacity respectively.
We again highlight an important change that has been integrated into the official 3DGS codebase of Kerbl~\etal~\cite{kerbl3Dgaussians} since the initial release is a change in the opacity learning rate, which was halved from 0.05 to 0.025 following Mallick~\etal~\cite{taming3dgs}.
As stated in the main paper, this change affects how many Gaussians are created during optimization with the new, lower learning rate leading to slightly cleaner reconstructions with fewer Gaussians.
For the Gaussian means, the learning rate is exponentially decayed from $1.6\text{e}\!-\!4$ to $1.6\text{e}\!-\!6$ during optimization and scaled by an additional constant factor that depends on the size of the scene to be reconstructed.
For the three view-independent components of the SH coefficients a learning rate of $2.5\text{e}\!-\!3$ is used, while remaining components use $1.25\text{e}\!-\!4$.
To prevent 3DGS from fitting diffuse color information with view-dependent SH coefficients, only the 0\textsuperscript{th} degree is used at the start of training and the next higher degree being enabled every 1000 iterations.

\section{Implementation Details}
In this section, we provide a detailed description of our refactored 3DGS implementation that is the \textit{basis} of our testbed as well details of our implementation for the improvements proposed in prior works (\cf \cref{ssec:improvements}).

\subsection{Testbed Basis}
\label{ssec:testbed_basis}
As our goal is to investigate the effectiveness of recently proposed optimizations for the original implementation by Kerbl~\etal~\cite{kerbl3Dgaussians}, we first start off by creating a clean, refactored 3DGS implementation as the basis for our testbed.
To avoid developing routines for data loading$\slash$management, logging, interactive viewing, and other functionalities commonly required for novel-view synthesis methods, we build our implementation on top on the NeRFICG framework~\cite{nerficg}.
A key advantage of it is that our \textit{Faster-GS} implementation retains modularity, simplifying future integration into other codebases.
Similar to Kerbl~\etal~\cite{kerbl3Dgaussians}, we use a PyTorch implementation with custom C++$\slash$CUDA extensions for frequently executed operations to ensure optimal performance.

The most important of these extensions is the differentiable software rasterizer for 3D Gaussians.
Our implementation is based on the original 3DGS, but features a complete rewrite of all CUDA kernels that includes various suitable simplifications and multiple small optimizations.
One of these simplifications involves removing the reliance on the OpenGL projection matrix where we instead use the intrinsic camera parameters directly (see \cref{eq:mean_projection,eq:cov_projection}).
This not only simplifies the math but also allows for properly rendering images with a non-centered optical center.
An arguably more significant change is that we compute the gradients of the alpha blending (see \cref{eq:alpha_blending}) in front-to-back order.
The original implementation does this in back-to-front order, which arguably makes the kernel code much more difficult to read.
More importantly, however, back-to-front order requires additional workarounds to ensure numerically stable gradient computation.
To this end, Kerbl~\etal first limit the maximum opacity a fragment can have when computing \cref{eq:alpha_computation} by computing $\hat{\alpha}=\min(0.99, \alpha)$, which ensures that repeated division by $(1 - \hat{\alpha_i})$ when computing the previous transmittance in the backward pass is stable.
For similar reasons, they use a somewhat unconventional approach for stopping the computation of \cref{eq:alpha_blending} early once the transmittance falls below a threshold $\tau=1\text{e}\!-\!4$.
Instead of first blending the fragment and checking whether to stop afterward, they first check whether the transmittance would be below $\tau$ if the fragment was to be blended and skip it if that check returns true.
In combination these fixes ensure numerical stability but arguably also create counterintuitive behavior in certain edge cases.
With our implementation using front-to-back order to compute the alpha blending gradients, we can get stable gradients without these workarounds.
In a similar spirit, we also employ a more direct approach for handling degenerate 3D Gaussians created by the parameter updates during training.
Specifically, a Gaussian in 3DGS is non-degenerate, if and only if its quaternion $q$ as well as all three of its scales are non-zero.
To obtain numerically stable gradients in single precision, these values must also not be too small.
Therefore, we do not render a Gaussian when $\|q\|<1\text{e}\!-\!4$ or $|\Sigma_\text{2D}|<1\text{e}\!-\!6$.
Note that as the former of these conditions is not view-dependent, we add it as an additional pruning condition during densification.

We also apply multiple small optimizations to the PyTorch-based frontend of our 3DGS framework.
First, we replace the implicit approach Kerbl~\etal use to make the rasterizer return the $\mu_\text{2D}$ gradients and visibility masks required as the metrics for densification by passing a persistent buffer storing these values to the rasterizer and marking it as non-differentiable.
This is slightly faster, requires less VRAM, and arguably much cleaner in general.
We also investigate VRAM fragmentation issues originating to frequent changes in buffer sizes during densification.
We find that, even when the maximum number of Gaussians is known before optimization, it will never be possible fully avoid these issues due to the number of tile instances being view- and optimization-dependent.
Fortunately, recent PyTorch versions support expandable memory segments, which enables lower and more predictable VRAM consumption during training.

Next, we revisit the implementation of the cloning, splitting, and pruning routines used for densification.
We find that the original implementation copies the Gaussians' parameters much more often than needed, which we avoid to eliminate unnecessary overhead.
Somewhat surprisingly, we also found that the official 3DGS implementation by Kerbl~\etal actually only performs 29855 optimizer steps with non-zero gradients because the densification routines zero out all gradients.
We fix this in our implementation by performing the optimizer step right after the backward pass.
While this means ours implementation performs 145 additional optimizer steps, it only makes a small difference in terms of runtime as the number of calls to the forward and backward passes of the rasterizer are not affected by this.

Lastly, we find that after a recent update the official 3DGS codebase clips the colors of the rendered images to the $[0, 1]$ range during training.
While the rendered colors can, due to the activation function applied to $c$, never take on negative values, values be larger than one are possible.
However, as clipping sets the gradient to zero, pixels with a final color marginally above one would not be used for optimization.
As this may cause unexpected behavior when the input images contain many white pixels due to, \eg, white walls or sky, we stick to the original approach of only clipping during inference.

\input{tables/x1_inference_rendering}

\subsection{Separate Sorting}
\label{ssec:separate_sorting_details}
As detailed in \cref{par:separate_sorting}, we follow Schütz~\etal~\cite{schuetz2025splatshop} and separate the sorting step for obtaining depth-sorted per-tile splat lists into two sorting routines, which has two key advantages.
First, we no longer need a 64-bit data type to store the keys for sorting leading to reduced VRAM usage.
The reason for this is that the first sort can use 32-bit keys for depth sorting while the second sort can use 16-bit (or 32-bit keys for very high image resolutions) for the tile sorting.
This is possible because the tiled rendering approach with 16\texttimes16 pixel tiles makes it so for images up to a resolution of $2^{16} \cdot 256$, \ie roughly 16 megapixels, tile indices fit into a 16-bit unsigned integer.
Note that this limit is slightly lower when image dimensions are not a multiple of 16.
The second advantage is in the complexity of the sorting, which for radix sort is $\mathcal{O}(kn)$ with k being the number of bits in the sorting key.
In the original 3DGS implementation, sorting is performed on a large buffer with 64-bit keys where each Gaussian can contribute an arbitrary amount of entries depending on the number of tiles it contributes to.
Importantly, the depth value in the lower 32 bits of the key is the same for all entries from the same Gaussian, while the tile index in the higher 32 bits will be different.
Assuming each of our $n$ Gaussians is visible in an average of 8 tiles, the complexity of the combined radix sort is $\mathcal{O}(8\cdot(16 + 32) n)\!=\!\mathcal{O}(384n)$
Note that we assume a key size of 48 bits as the radix sort implementation used by Kerbl~\etal supports accounting for the fact that not all of the higher 32 bits are needed to store the tile indices (usually 16 bits suffice) and can therefore be disregarded during sorting.
When separating the sorting, the complexities of the two steps become $\mathcal{O}(32n)$ and $\mathcal{O}(8\cdot16 n)$ respectively, \ie the combined complexity is $\mathcal{O}(32n) + \mathcal{O}(128n)$.
This clearly shows the advantage of using the separate sorting approach with increasingly higher benefits as the number of Gaussians increases.

Similar to Schütz~\etal~\cite{schuetz2025splatshop}, we identify the indices of all visible Gaussians as well as their depth during pre-processing and write them to compacted buffers used for depth sorting through the use of an atomic counter.
An important change we make to optimize training performance is that we do not apply this compaction to the intermediate Gaussian values needed for rasterization and blending.
This allows our implementation to avoid an additional indirection during gradient computation in the backward pass.

\subsection{Per-Gaussian Backward}
\label{ssec:per_gaussian_backward_details}
A major part of the speedup in our optimized implementation comes from the per-Gaussian backward pass (\cref{par:per_gaussian_backward}) proposed by Mallick~\etal~\cite{taming3dgs}.
It requires three major changes to the implementation.
First, additional buffers are allocated to store the intermediate color and transmittance after every 32\textsuperscript{nd} Gaussian as well as the final number of contributing Gaussians at each pixel.
Note that the size of these buffers can easily be determined from the known lengths of the per-tile lists.
Second, the forward pass for blending needs to fill these buffers with the corresponding values, which introduces negligible overhead.
Third, the backward pass for blending now operates on so-called buckets of 32 Gaussians each where each bucket is associated with a single tile.
For each bucket, 32 threads (\ie, one warp) are launched with each thread accumulating the gradient of a single Gaussian across the tile associated with the respective bucket.
Starting from the buckets initial color and transmittance at each pixel as written in the forward pass, the threads replay the blending process using warp-level primitives to efficiently compute the necessary gradients.
Finally, each thread writes the accumulated gradients to global memory using atomics as Gaussians can be inside multiple buckets across tiles. 
We refer the reader to the the original paper for further details~\cite{taming3dgs}.

In addition to integrating our changes with respect to early stopping and numerical stability (see \cref{ssec:basis_impl}), we also extend the idea of Mallick~\etal for improved performance.
In their implementation, the first thread in each warp repeatedly loads the alpha blending state for the next pixel from global memory.
As threads must remain synchronized, the entire warp stalls on these global loads.
We optimize this by having all 32 threads in the warp collaboratively load a batch of alpha blending states into shared memory (one per thread) before they are needed.
From that point on, the first thread in each warp reads the next state directly from shared instead of global memory.
Because shared memory access has much lower latency than global memory, this significantly reduces warp stalls.
Profiling shows that this change improves the kernel runtime by up to 2\texttimes.

\section{Further Experiments and Results}
In the following, we present results of multiple experiments that are complementary to the evaluation in the main paper.

\subsection{Inference Rendering Performance}
\label{ssec:inference_fps}
Apart from rapid optimization, another highly relevant aspect of efficient 3D Gaussian Splatting is fast inference rendering.

Of course, many of the improvements that we implement in our testbed (\cref{sec:method}) positively influence frame rate.
In \cref{tab:inference_fps}, we show that an inference-optimized version of the forward pass from our testbed leads to more than 3\texttimes\ faster rendering  during inference.

\subsection{Efficient Anti-Aliasing}
\label{ssec:aa_results}
In Mip-Splatting~\cite{yu2024mip}, Yu~\etal propose two extensions for optimization and rendering that, in combination, effectively prevent aliasing artifacts that occur in the original 3DGS approach when changing the sampling rate after training, \eg, by adjusting the focal length or camera distance.

The first extension is a 3D smoothing filter that prevents Gaussians from becoming smaller than the maximal sampling frequency induced by the images used for training.
For each primitive, they define the maximal sampling rate of the $k$\textsuperscript{th} Gaussian as
\begin{equation}
\hat{\nu}_k = \max \left(\left\{ \mathds{1}_n(\mathbf{\mu}_k) \cdot \frac{f_n}{d_n}\right\}^{N}_{n=1}\right),
\end{equation}
where $\mathds{1}_n(\cdot)$ is an indicator function stating whether the input point is visible in the $n$\textsuperscript{th} training image, $N$ is the number of training images, $f_n$ is the focal length of the $n$\textsuperscript{th} training view in pixel units, and $d_n$ is the $z$-depth of $\mathbf{\mu}_k$ for the respective view.
As $\hat{\nu}_k$ changes whenever $\mathbf{\mu}_k$ is updated during optimization, Yu~\etal recompute this value for all Gaussians every 100 iterations during training.

For rendering, the 3D smoothing filter is applied through a Gaussian low-pass filter that influences the three scales $\mathbf{s}_k$ and the opacity $o_k$ of each Gaussian:
\begin{equation}\label{eq:3dfilter_application}
\hat{\mathbf{s}}_k = \sqrt{\mathbf{s}_k^2 + \frac{\kappa_\text{3D}}{\hat{\nu}_k^2}} \quad \text{and} \quad
\hat{o}_k = \sqrt{\frac{|\text{diag}(\mathbf{s}_k^2)|}{|\text{diag}(\mathbf{s}_k^2 + \frac{\kappa_\text{3D}}{\hat{\nu}_k^2})|}} o_k,
\end{equation}
where $\kappa_\text{3D}$ is a hyperparameter controlling the variance of the Gaussian filter ($0.2$ by default).

While disabled by default, we implement two version of this 3D smoothing filter in \textit{Faster-GS}.
The first version closely matches the original implementation by Yu~\etal.
For the second version, we closely adhere to the underlying theory but use a more direct and efficient approach for enforcing the implied size constraints during optimization.
Specifically, we in-place clip the scales of each Gaussian from below based on the 3D smoothing filter after each optimizer step:
\begin{equation}
\mathbf{s}_k = \max(\mathbf{s}_k, \frac{\sqrt{\kappa_\text{3D}}}{\hat{\nu}_k}).
\end{equation}
We find that this retains all advantages while being significantly cheaper to compute as this update is independent of gradient computations.
Furthermore, we find that accounting for the change in volume of each Gaussian by modifying its opacity is not needed in practice (\cf Steiner~\etal~\cite{steiner2025aaags}), allowing for further simplifications.
To accelerate the repeated computation of $\hat{\nu}_k$ during training, we use a fused CUDA implementation by Hahlbohm~\etal~\cite{hahlbohm2025htgs}.

The second extension of Yu~\etal is a 2D Mip filter that mitigates artifacts when rendering Gaussians from further away or with larger focal length than during training.
It is an extension to the 2D Gaussian filter that Kerbl~\etal use in the original 3DGS~\cite{kerbl3Dgaussians} to prevent aliasing caused by projected 2D Gaussians falling between the pixels due to being too small.
Kerbl~\etal use a 2D Gaussian filter with variance $\kappa_\text{3D}$ ($0.3$ by default):
\begin{equation}\label{eq:original_2dfilter}
\hat{\Sigma}_\text{2D} = \Sigma_\text{2D} + \kappa_\text{3D}\text{I}.
\end{equation}
While this approach of dilating every Gaussian works very well during optimization, it does cause aliasing issues during inference.
As the virtual camera moves further from a Gaussian it becomes smaller and smaller until the point where $\hat{\Sigma}_\text{2D}$ in \cref{eq:original_2dfilter} is dominated by the dilation kernel, which leads to increasingly blurry renderings.
To avoid this, Yu~\etal multiply a view-dependent compensation factor onto the opacity of each Gaussian:
\begin{equation}\label{eq:filter2d_compensation}
\hat{o} = \sqrt{\frac{|\Sigma_\text{2D}|}{|\hat{\Sigma}_\text{2D}|}} o.
\end{equation}
Note that because the full approach of Yu~\etal combines the 3D smoothing filter with this 2D Mip filter, they use a smaller variance for the 2D Gaussian filter ($0.1$ by default).

\input{tables/x2_anti_aliasing}
\input{figures/x1_anti_aliasing}

To complement the 3D smoothing filter, we also integrate the 2D Mip filter into \textit{Faster-GS}.
For optimal performance, we make sure to use the smaller opacity values resulting from \cref{eq:filter2d_compensation} to when computing opacity-aware bounding boxes (see \cref{par:splat_bounding}).
We also find that the derivatives of \cref{eq:filter2d_compensation} \wrt $\Sigma_\text{2D}$ can be numerically unstable causing gradients to explode.
When investigating this, we found that the original implementation~\cite{yu2024mip} frequently clips extreme values, while re-implementations~\cite{taming3dgs, radl2024stopthepop} compute gradients in a way that does not match the analytical derivative and it is unclear whether this is done on purpose.
We find that a much more effective and practical approach to this issue is to simply detach the compensation factor \cref{eq:filter2d_compensation} from the gradient computation for $\Sigma_\text{2D}$.
Note that we still provide a reasonably stable implementation for the full analytical derivative that can optionally be enabled.

\input{tables/x3_multi_scale_evaluation}

In \cref{tab:anti_aliasing}, we show results for a single-scale training and same-scale evaluation experiment on the nine scenes from the Mip-NeRF360 dataset~\cite{barron2022mipnerf360}.
Our simplifications for the 3D smoothing filter effectively eliminate any training overhead compared to the original implementation~\cite{yu2024mip}.
We also find that our changes to the backward pass of the 2D Mip filter result in more stable optimization behavior as indicated by the average number of Gaussians that is more similar to the baseline.
Specifically, in the original implementation as well as in \textit{Faster-GS} with only the 2D Mip filter enabled, we find that the optimization sometimes creates additional tiny and elongated, \ie, degenerate Gaussians, which slightly reduce overall quality of the reconstruction.
Most importantly, however, we find that our optimized implementation of the two extensions from Yu~\etal~\cite{yu2024mip} enables fully anti-aliased training and rendering inside our framework.
Note that we adjusted the official Mip-Splatting implementation to use the fused SSIM implementation from Taming-3DGS~\cite{taming3dgs} and the updated opacity learning rate for fair comparison (\cf \cref{ssec:experiment_setup}).
We further validate the effectiveness of the implemented anti-aliasing approach in \cref{tab:multi_scale_eval} and \cref{fig:aa_grid}.

\subsection{Fast MCMC Densification}
\label{ssec:mcmc_results}
In their work 3DGS-MCMC, Kheradmand~\etal~\cite{kheradmand243dgsmcmc} propose an alternative approach for densification that treats the set of 3D Gaussians as Markov Chain Monte Carlo (MCMC) samples.
Based on Stochastic Gradient Langevin Dynamics (SGLD) updates, they add noise to the Gaussian means after each training iteration.
The splitting, cloning, and pruning steps used in adaptive density control~\cite{kerbl3Dgaussians} are replaced by a re-localization scheme that tries to preserves sample probability.
Gaussian are selected for re-localization when they can no longer meaningfully contribute to renderings, \ie, when they are small or have low opacity.
To encourage optimal distribution of a preset number of Gaussians, Kheradmand~\etal also add regularization terms to the loss function.

This approach to 3DGS densification has three main advantages: it reduces reliance on the initial point cloud, allows for specifying the number of Gaussians prior to optimization, and it leads to improved rendering quality.
These advantages motivate us to integrate an optimized version into our \textit{Faster-GS} framework.
Specifically, we fuse the noise injection step into a single CUDA kernel, as we found that it is a main bottleneck \wrt training time.
In \cref{tab:mcmc}, we compare our optimized version with the original implementation~\cite{kheradmand243dgsmcmc} on the nine scenes from the Mip-NeRF360 dataset~\cite{barron2022mipnerf360}.
Note that we adjusted the original implementation of 3DGS-MCMC~\cite{kheradmand243dgsmcmc} to use the fused SSIM implementation from Taming-3DGS~\cite{taming3dgs} for fair comparison.

For the sake of reproducibility, we provide the target number of Gaussians for the nine scenes in alphabetical order: 6131954, 1244819, 1222956, 3636448, 5834784, 1852335, 1593376, 4961797, 3783761 (values taken from Steiner~\etal~\cite{steiner2025aaags}).

\input{tables/x4_mcmc}

\subsection{About Gaussian Truncation and Opacity}
In this subsection, we will investigate the effects of truncating Gaussians at different standard deviations as well as an alternative interpretation of the Gaussian opacities in 3DGS.

By definition, Gaussians have infinite support, \ie, have a non-zero value at any query point.
For rendering a Gaussian, however, the near-zero values obtained when querying it far away from its mean can safely be skipped.
Therefore, the original 3DGS approach~\cite{kerbl3Dgaussians} truncates the projected 2D Gaussians at roughly $3.33\sigma$.
Importantly, Kerbl~\etal do not solely truncate based on the standard deviation of the 2D Gaussian as defined by $\Sigma_\text{2D}$ (see \cref{eq:cov_projection}) but also factor in opacity.
In their implementation, they achieve this by skipping all fragments during blending where the computed transparency value $\alpha$ (see \cref{eq:alpha_computation}) is below a threshold $\tau_\alpha\!=\!\nicefrac{1}{255}$.
However, because computing $\alpha$ involves the opacity, a Gaussian with an opacity below $\tau_\alpha$ can never be visible or receive gradients in the original implementation.
Therefore, the clipping value of $\nicefrac{1}{100}$ used by the opacity reset enforces an upper bound for $\tau_\alpha$.
A larger threshold would result in all fragments being discarded after the first reset.
This limit results in truncation at $\sqrt{-2\ln(\nicefrac{1}{100})}\approx\!3.03$ standard deviations for Gaussians that have an opacity close to one.
In other words, the approach for implicit Gaussian truncation based on fragment $\alpha$ used in the original 3DGS implementation prevents consistent, opacity-independent truncation at fewer than $3.03$ standard deviations during training.

\input{tables/x5_truncation}

We propose a modification that can avoid this issue.
Our idea is to check whether the response of a Gaussian at a given pixel is below $\tau_\alpha$ before multiplying by the opacity.
This then allows for truncation at fewer standard deviations and also addresses an issue \wrt how 3DGS computes opacity gradients in the backward pass that was recently brought up by Hahlbohm~\etal~\cite{hahlbohm2025inpcv2}:
The analytical derivatives of \cref{eq:alpha_computation,eq:alpha_blending} provide a non-zero gradient for the opacity of a Gaussian even when its value is zero.
In the original implementation, Kerbl~\etal disregard these gradients as they interpret the opacity as part of the Gaussian response.
While this is a reasonable approach that clearly works well in practice, we still think it is worth investigating.
We also think that more aggressive truncation that uses, \eg, 2$\sigma$ could be an interesting avenue for future work that renders splats as opaque 2D ellipses to avoid the need for depth-ordered alpha blending.

We show results for different truncation configurations in \cref{tab:truncation}.
Note that all version use our full anti-aliasing as we find that it integrates particularly well with the opacity-independent approach for truncation.

\subsection{Additional 4D Reconstruction Results}
\label{ssec:n3dv_results}
We also tested our extension to 4D Gaussians on three real scenes from the multi-view dataset by Li~\etal~\cite{li2022neural3dv}.
We preprocess scenes similar to Yang~\etal to obtain an initial point cloud and use a consistent hyperparameter configuration for their baseline~\cite{yang20244dgaussians} and our implementation:
Scenes are trained and evaluated at 1352\texttimes1014 using the provided train$\slash$test splits, models use the standard view-dependent color parametrization from 3DGS (SH up to degree three), and training is done for 30000 iterations with a batch size of four.
Results are shown in \cref{tab:n3dv_results}.
\input{tables/x6_n3dv_results}

%% file: tables/x1_inference_rendering.tex
\begin{table*}[t]
\caption{%
Inference frame rates for all 13 scenes from the Mip-NeRF360, Tanks and Temples, and Deep Blending datasets~\cite{barron2022mipnerf360, Knapitsch2017, hedman2018deep}. The depicted values are the average frames per second when rendering the test set of the respective scene 100 times at the native resolution. For benchmarking 3DGS, we follow Hahlbohm~\etal~\cite{hahlbohm2025htgs} and bake all activation functions before rendering to avoid any PyTorch-related overhead. For Ours, we add an inference-optimized version of the forward pass to our testbed where we enable all improvements (\cf \cref{sec:method}) that accelerate inference.
}\label{tab:inference_fps}
\centering
\setlength\tabcolsep{2.2pt}
\small
\begin{tabular}{lccccccccccccc|c}
\toprule
 & Bicycle & Flowers & Garden & Stump & Treehill & Bonsai & Counter & Kitchen & Room & Train & Truck & DrJohnson & Playroom & Average \\
\midrule
3DGS~\cite{kerbl3Dgaussians} & 161.8 & 387.8 & 223.7 & 329.7 & 306.2 & 405.4 & 261.0 & 219.6 & 253.2 & 317.5 & 340.8 & 221.3 & 348.0 & 290.4 \\
Ours                         & 547.1 & 901.3 & 628.5 & 821.0 & 824.7 & 1239.7 & 1018.3 & 745.4 & 1122.4 & 833.0 & 863.1 & 919.2 & 1280.8 & 903.4 \\
~$\hookrightarrow$ speedup   & 3.4\texttimes & 2.3\texttimes & 2.8\texttimes & 2.5\texttimes & 2.7\texttimes & 3.1\texttimes & 3.9\texttimes & 3.4\texttimes & 4.4\texttimes & 2.6\texttimes & 2.5\texttimes & 4.2\texttimes & 3.7\texttimes & 3.1\texttimes \\
\bottomrule
\end{tabular}
\end{table*}

%% file: tables/x2_anti_aliasing.tex
\begin{table}[t]
\caption{%
Quantitative comparisons of approaches for anti-aliasing based on Mip-Splatting on the nine scenes from the Mip-NeRF360 dataset~\cite{barron2022mipnerf360}.
All approaches optimize to the same quality, but the anti-aliased version of \textit{Faster-GS} trains much faster and requires less VRAM compared to the original implementation~\cite{yu2024mip}. We highlight the significant speedup from our simplified 3D smoothing filter as well as the a more consistent number of Gaussians due to our revised backward pass for the 2D Mip filter.
}\label{tab:anti_aliasing}
\centering
\setlength\tabcolsep{5.4pt}
\small
\begin{tabular}{lcccc}
\toprule
 & PSNR$^\uparrow$ & Train$^\downarrow$ & VRAM$^\downarrow$ & \#Gs$^\downarrow$ \\
\midrule
Mip-Splatting~\cite{yu2024mip} & 27.53 &  19m56s & 8.0GiB  & 2.82M \\
Ours                           & 27.56 & ~~4m31s & 6.1GiB  & 2.73M \\
~~+ original 3D filter         & 27.53 & ~~5m58s & 6.3GiB  & 2.71M \\
~~+ our 3D filter              & 27.55 & ~~4m32s & 6.1GiB  & 2.72M \\
~~+ 2D Mip filter              & 27.54 & ~~4m30s & 6.2GiB  & 2.78M \\
~~+ full anti-aliasing         & 27.54 & ~~4m31s & 6.1GiB  & 2.70M \\
\bottomrule
\end{tabular}
\end{table}

%% file: figures/x1_anti_aliasing.tex
\begin{figure*}[tb]
\centering
\setlength\tabcolsep{0pt}
\begin{tabular}{*{3}{p{0.33333333333333\linewidth}<{\centering}}@{}}
1\texttimes\ Resolution & 0.5\texttimes\ Resolution & 2\texttimes\ Resolution
\end{tabular}
\includegraphics[width=\linewidth]{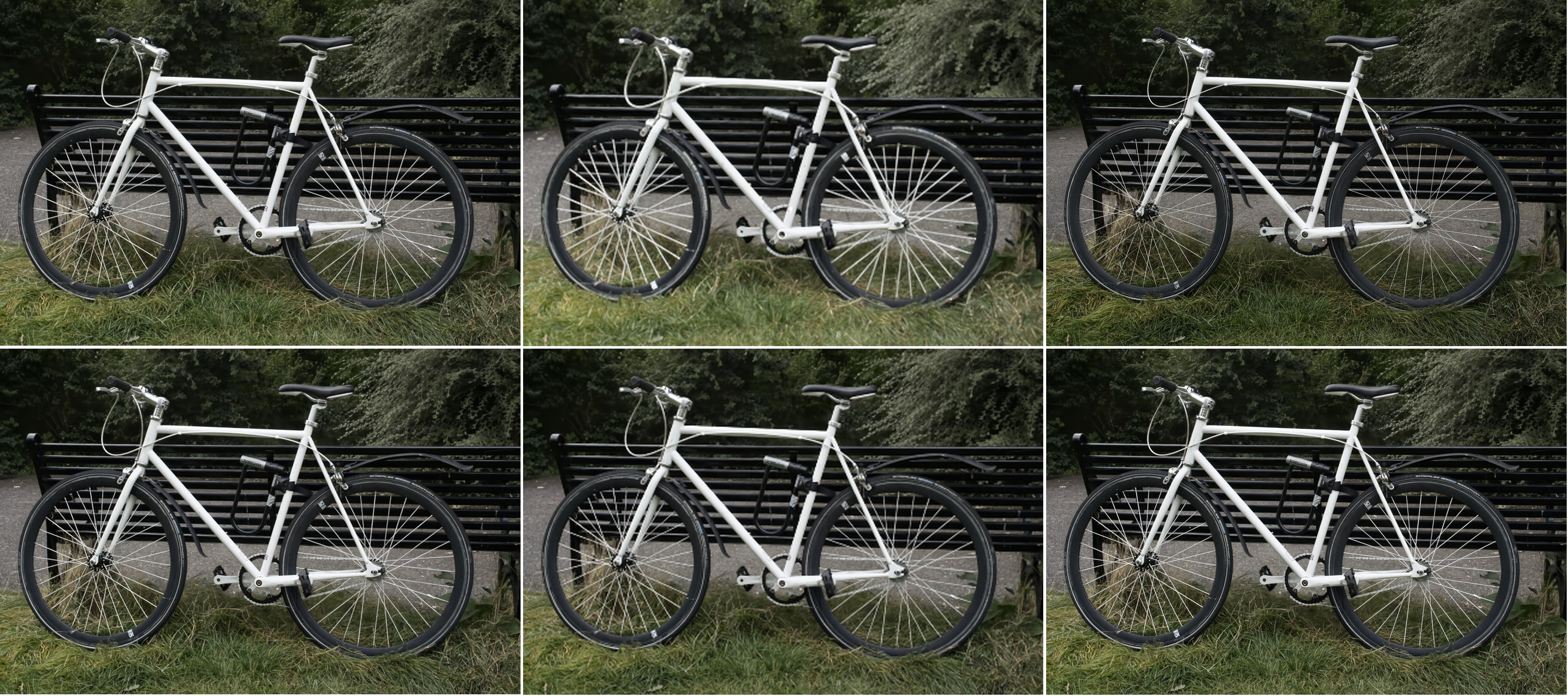}
\caption{%
We show renderings of two of our models. The first one (top) was trained without anti-aliasing techniques, the second one (bottom) has them enabled. It is clearly visible that rendering at a different resolution compared to the one used during training (1\texttimes) leads to aliasing artifacts (top). The implemented anti-aliasing techniques significantly reduce these artifacts leading to higher visual fidelity (bottom).
}\label{fig:aa_grid}
\end{figure*}

%% file: tables/x3_multi_scale_evaluation.tex
\begin{table}[tb]
\caption{%
Single-scale training and multi-scale evaluation on the nine scenes from the Mip-NeRF360 dataset~\cite{barron2022mipnerf360}. For both approaches, training is done at the default scene resolution, \ie, with 4\texttimes$\slash$2\texttimes\ downsampling for outdoor$\slash$indoor scenes respectively. We then evaluate the model at the training resolution (1\texttimes) as well as at half (0.5\texttimes) and double (2\texttimes) that resolution for each scene. The results confirm the effectiveness of the anti-aliased version of \textit{Faster-GS}.
}\label{tab:multi_scale_eval}
\centering
\setlength\tabcolsep{9.6pt}
\small
\begin{tabular}{lccc}
\toprule
 & \multicolumn{3}{c}{PSNR$^\uparrow$} \\
 & 1\texttimes\ Res. & 0.5\texttimes\ Res. & 2\texttimes\ Res. \\
\midrule
Ours                           & 27.56 & 25.11 & 25.73 \\
~~+ full anti-aliasing         & 27.54 & 28.39 & 26.87 \\
\bottomrule
\end{tabular}
\end{table}

%% file: tables/x4_mcmc.tex
\begin{table}[b]
\caption{%
Quantitative comparison of our and the original implementation~\cite{kheradmand243dgsmcmc} for MCMC densification on the nine scenes from the Mip-NeRF360 dataset~\cite{barron2022mipnerf360}. While both versions achieve similar quality, our optimized implementation is significantly faster and uses less VRAM during training.
}\label{tab:mcmc}
\centering
\setlength\tabcolsep{10pt}
\small
\begin{tabular}{lccc}
\toprule
 & PSNR$^\uparrow$ & Train$^\downarrow$ & VRAM$^\downarrow$ \\
\midrule
3DGS-MCMC~\cite{kheradmand243dgsmcmc} & 27.83 &  27m22s & 8.9GiB \\
Ours                                  & 28.00 & ~~6m17s & 7.2GiB \\
\bottomrule
\end{tabular}
\end{table}

%% file: tables/x5_truncation.tex
\begin{table}[b]
\caption{%
Quantitative comparisons of truncating Gaussians at different standard deviations on the nine scenes from the Mip-NeRF360 dataset~\cite{barron2022mipnerf360}. We find that densification creates fewer Gaussians when truncating at 1$\slash$2$\sigma$ leading to reduced quality. While our modification for opacity-independent truncation slows down training, we think that the increased flexibility provides an interesting avenue for future work. We highlight that the most aggressive truncation (1$\sigma$) makes it so the minimum opacity of a contributing fragment is $\approx\!0.61$, which means Gaussians are close to opaque. All configurations use our full anti-aliasing (see \cref{tab:anti_aliasing}).
}\label{tab:truncation}
\centering
\setlength\tabcolsep{4.9pt}
\small
\begin{tabular}{lcccc}
\toprule
 & PSNR$^\uparrow$ & Train$^\downarrow$ & VRAM$^\downarrow$ & \#Gs$^\downarrow$ \\
\midrule
3.33$\sigma$ (default)              & 27.54 & 4m31s & 6.1GiB & 2.70M \\
4$\sigma$                           & 27.54 & 5m05s & 6.4GiB & 2.78M \\
4$\sigma$ w$\slash$ modification    & 27.65 & 5m27s & 6.6GiB & 2.78M \\
3.33$\sigma$ w$\slash$ modification & 27.57 & 5m07s & 6.4GiB & 2.79M \\
3$\sigma$ w$\slash$ modification    & 27.62 & 4m55s & 6.3GiB & 2.78M \\
2$\sigma$ w$\slash$ modification    & 27.23 & 3m41s & 5.5GiB & 2.22M \\
1$\sigma$ w$\slash$ modification    & 25.86 & 2m26s & 4.5GiB & 1.39M \\
\bottomrule
\end{tabular}
\end{table}

%% file: tables/x6_n3dv_results.tex
\begin{table}[h]
\caption{%
Comparison with the reference implementation~\cite{yang20244dgaussians} for our extension to 4D Gaussians on three scenes (\textit{Coffee Martini}, \textit{Cook Spinach}, and \textit{Flame Steak}) from the neural 3D video dataset~\cite{li2022neural3dv}.
}\label{tab:n3dv_results}
\centering
\setlength\tabcolsep{11.8pt}
\small
\begin{tabular}{lccc}
\toprule
 & PSNR$^\uparrow$ & Train$^\downarrow$ & VRAM$^\downarrow$ \\
\midrule
Yang~\etal~\cite{yang20244dgaussians} & 30.13 & 96m43s & 8.7GiB \\
Ours                                  & 30.54 & 18m57s & 3.7GiB \\
\bottomrule
\end{tabular}
\end{table}

%% file: main.bbl
\begin{thebibliography}{103}
\providecommand{\natexlab}[1]{#1}
\providecommand{\url}[1]{\texttt{#1}}
\expandafter\ifx\csname urlstyle\endcsname\relax
  \providecommand{\doi}[1]{doi: #1}\else
  \providecommand{\doi}{doi: \begingroup \urlstyle{rm}\Url}\fi

\bibitem[Bagdasarian et~al.(2025)Bagdasarian, Knoll, Li, Barthel, Hilsmann,
  Eisert, and Morgenstern]{3DGSzip2024}
Milena~T. Bagdasarian, Paul Knoll, Yi-Hsin Li, Florian Barthel, Anna Hilsmann,
  Peter Eisert, and Wieland Morgenstern.
\newblock {3DGS.zip}: A survey on {3D Gaussian} splatting compression methods.
\newblock \emph{CGF}, 44\penalty0 (2), 2025.

\bibitem[Barron et~al.(2022)Barron, Mildenhall, Verbin, Srinivasan, and
  Hedman]{barron2022mipnerf360}
Jonathan~T. Barron, Ben Mildenhall, Dor Verbin, Pratul~P. Srinivasan, and Peter
  Hedman.
\newblock {Mip-NeRF} 360: Unbounded anti-aliased neural radiance fields.
\newblock In \emph{CVPR}, pages 5460--5469, 2022.

\bibitem[Blanc et~al.(2025{\natexlab{a}})Blanc, Deschaud, and
  Paljic]{blanc2025raygauss}
Hugo Blanc, Jean-Emmanuel Deschaud, and Alexis Paljic.
\newblock {RayGauss}: Volumetric {Gaussian}-based ray casting for
  photorealistic novel view synthesis.
\newblock In \emph{WACV}, 2025{\natexlab{a}}.

\bibitem[Blanc et~al.(2025{\natexlab{b}})Blanc, Deschaud, and
  Paljic]{blanc2025raygaussx}
Hugo Blanc, Jean-Emmanuel Deschaud, and Alexis Paljic.
\newblock {RayGaussX}: Accelerating {Gaussian}-based ray marching for real-time
  and high-quality novel view synthesis.
\newblock In \emph{ICCV}, 2025{\natexlab{b}}.

\bibitem[Celarek et~al.(2025)Celarek, Kopanas, Drettakis, Wimmer, and
  Kerbl]{celarek2025does3dgsneedvolrend}
Adam Celarek, Georgios Kopanas, George Drettakis, Michael Wimmer, and Bernhard
  Kerbl.
\newblock Does 3d gaussian splatting need accurate volumetric rendering?
\newblock \emph{CGF}, 44\penalty0 (2), 2025.

\bibitem[Charatan et~al.(2024)Charatan, Li, Tagliasacchi, and
  Sitzmann]{charatan2024pixelsplat}
David Charatan, Sizhe~Lester Li, Andrea Tagliasacchi, and Vincent Sitzmann.
\newblock {PixelSplat}: {3D Gaussian} splats from image pairs for scalable
  generalizable {3D} reconstruction.
\newblock In \emph{CVPR}, pages 19457--19467, 2024.

\bibitem[Chen et~al.(2025{\natexlab{a}})Chen, Li, Ye, Wang, Xie, Zhai, Wang,
  Liu, Bao, and Zhang]{chen2025pgsr}
Danpeng Chen, Hai Li, Weicai Ye, Yifan Wang, Weijian Xie, Shangjin Zhai, Nan
  Wang, Haomin Liu, Hujun Bao, and Guofeng Zhang.
\newblock {PGSR}: Planar-based {Gaussian} splatting for efficient and
  high-fidelity surface reconstruction.
\newblock \emph{IEEE TVCG}, 31\penalty0 (9):\penalty0 6100--6111,
  2025{\natexlab{a}}.

\bibitem[Chen et~al.(2024)Chen, Chen, Zhang, Wang, Yang, Wang, Cai, Yang, Liu,
  and Lin]{chen2024gaussianeditor}
Yiwen Chen, Zilong Chen, Chi Zhang, Feng Wang, Xiaofeng Yang, Yikai Wang,
  Zhongang Cai, Lei Yang, Huaping Liu, and Guosheng Lin.
\newblock {GaussianEditor}: Swift and controllable {3D} editing with {Gaussian}
  splatting.
\newblock In \emph{CVPR}, pages 21476--21485, 2024.

\bibitem[Chen et~al.(2025{\natexlab{b}})Chen, Jiang, Jiang, Tang, Li, Liu, and
  Nie]{chen2025dashgaussian}
Youyu Chen, Junjun Jiang, Kui Jiang, Xiao Tang, Zhihao Li, Xianming Liu, and
  Yinyu Nie.
\newblock Dashgaussian: Optimizing 3d gaussian splatting in 200 seconds.
\newblock In \emph{CVPR}, 2025{\natexlab{b}}.

\bibitem[Chen et~al.(2025{\natexlab{c}})Chen, Xu, Zheng, Zhuang, Pollefeys,
  Geiger, Cham, and Cai]{chen2025mvsplat}
Yuedong Chen, Haofei Xu, Chuanxia Zheng, Bohan Zhuang, Marc Pollefeys, Andreas
  Geiger, Tat-Jen Cham, and Jianfei Cai.
\newblock {MVSplat}: Efficient {3D Gaussian} splatting from sparse multi-view
  images.
\newblock In \emph{ECCV}, pages 370--386, 2025{\natexlab{c}}.

\bibitem[Chung et~al.(2024)Chung, Oh, and Lee]{chung2024depthreg3dgs}
Jaeyoung Chung, Jeongtaek Oh, and Kyoung~Mu Lee.
\newblock Depth-regularized optimization for {3D Gaussian} splatting in
  few-shot images.
\newblock In \emph{CVPRW}, 2024.

\bibitem[Condor et~al.(2025)Condor, Speierer, Bode, Bozic, Green, Didyk, and
  Jarabo]{condor2025dontsplat}
Jorge Condor, Sebastien Speierer, Lukas Bode, Aljaz Bozic, Simon Green, Piotr
  Didyk, and Adrian Jarabo.
\newblock Don't splat your {Gaussians}: Volumetric ray-traced primitives for
  modeling and rendering scattering and emissive media.
\newblock \emph{ACM TOG}, 44\penalty0 (1), 2025.

\bibitem[Durvasula et~al.(2023)Durvasula, Zhao, Chen, Liang, Sanjaya, and
  Vijaykumar]{durvasula2023distwar}
Sankeerth Durvasula, Adrian Zhao, Fan Chen, Ruofan Liang, Pawan~Kumar Sanjaya,
  and Nandita Vijaykumar.
\newblock Distwar: Fast differentiable rendering on raster-based rendering
  pipelines, 2023.

\bibitem[Fang and Wang(2024)]{fang2025minisplatting}
Guangchi Fang and Bing Wang.
\newblock {Mini-Splatting}: Representing scenes with a constrained number of
  {Gaussians}.
\newblock In \emph{ECCV}, pages 165--181, 2024.

\bibitem[Feng et~al.(2025)Feng, Chen, Fu, Liao, Wang, Liu, Hu, Xu, Pei, Li, Li,
  Sun, Zhang, and Dai]{feng2025flashgs}
Guofeng Feng, Siyan Chen, Rong Fu, Zimu Liao, Yi Wang, Tao Liu, Boni Hu, Lining
  Xu, Zhilin Pei, Hengjie Li, Xiuhong Li, Ninghui Sun, Xingcheng Zhang, and Bo
  Dai.
\newblock {FlashGS}: Efficient {3D Gaussian} splatting for large-scale and
  high-resolution rendering.
\newblock In \emph{CVPR}, pages 26652--26662, 2025.

\bibitem[Franke et~al.(2024)Franke, R\"{u}ckert, Fink, and
  Stamminger]{franke2024trips}
Linus Franke, Darius R\"{u}ckert, Laura Fink, and Marc Stamminger.
\newblock {TRIPS}: Trilinear point splatting for real-time radiance field
  rendering.
\newblock \emph{CGF}, 43\penalty0 (2), 2024.

\bibitem[Franke et~al.(2025)Franke, Fink, and
  Stamminger]{franke2025vrsplatting}
Linus Franke, Laura Fink, and Marc Stamminger.
\newblock {VR-Splatting}: Foveated radiance field rendering via {3D Gaussian}
  splatting and neural points.
\newblock \emph{PACMCGIT}, 8\penalty0 (1), 2025.

\bibitem[Girish et~al.(2024)Girish, Gupta, and Shrivastava]{girish2024eagles}
Sharath Girish, Kamal Gupta, and Abhinav Shrivastava.
\newblock {EAGLES}: Efficient accelerated {3D Gaussians} with lightweight
  encodings.
\newblock In \emph{ECCV}, pages 54--71, 2024.

\bibitem[Grubert et~al.(2025)Grubert, Barthel, Hilsmann, and
  Eisert]{grubert2025improvedadc}
Glenn Grubert, Florian Barthel, Anna Hilsmann, and Peter Eisert.
\newblock Improving adaptive density control for 3d gaussian splatting.
\newblock In \emph{VISIGRAPP}, 2025.

\bibitem[Gu{\'e}don and Lepetit(2024)]{guedon2024sugar}
Antoine Gu{\'e}don and Vincent Lepetit.
\newblock {SuGaR}: Surface-aligned gaussian splatting for efficient 3d mesh
  reconstruction and high-quality mesh rendering.
\newblock In \emph{CVPR}, pages 5354--5363, 2024.

\bibitem[Gui et~al.(2024)Gui, Hu, Chen, Huang, Yin, Yang, Wu, Liu, Sun, Zhang,
  et~al.]{gui2024balanced3dgs}
Hao Gui, Lin Hu, Rui Chen, Mingxiao Huang, Yuxin Yin, Jin Yang, Yong Wu, Chen
  Liu, Zhongxu Sun, Xueyang Zhang, et~al.
\newblock {Balanced 3DGS}: Gaussian-wise parallelism rendering with
  fine-grained tiling, 2024.

\bibitem[Guédon et~al.(2025)Guédon, Gomez, Maruani, Gong, Drettakis, and
  Ovsjanikov]{guedon2025milo}
Antoine Guédon, Diego Gomez, Nissim Maruani, Bingchen Gong, George Drettakis,
  and Maks Ovsjanikov.
\newblock {MILo}: Mesh-in-the-loop {Gaussian} splatting for detailed and
  efficient surface reconstruction.
\newblock \emph{ACM TOG}, 2025.

\bibitem[Hahlbohm et~al.(2025{\natexlab{a}})Hahlbohm, Franke, Overkämping,
  Wespe, Castillo, Eisemann, and Magnor]{hahlbohm2025inpcv2}
Florian Hahlbohm, Linus Franke, Leon Overkämping, Paula Wespe, Susana
  Castillo, Martin Eisemann, and Marcus Magnor.
\newblock A bag of tricks for efficient implicit neural point clouds.
\newblock In \emph{VMV}, 2025{\natexlab{a}}.

\bibitem[Hahlbohm et~al.(2025{\natexlab{b}})Hahlbohm, Friederichs, Weyrich,
  Franke, Kappel, Castillo, Stamminger, Eisemann, and Magnor]{hahlbohm2025htgs}
Florian Hahlbohm, Fabian Friederichs, Tim Weyrich, Linus Franke, Moritz Kappel,
  Susana Castillo, Marc Stamminger, Martin Eisemann, and Marcus Magnor.
\newblock Efficient perspective-correct {3D Gaussian} splatting using hybrid
  transparency.
\newblock \emph{CGF}, 44\penalty0 (2), 2025{\natexlab{b}}.

\bibitem[Hanson et~al.(2025)Hanson, Tu, Lin, Singla, Zwicker, and
  Goldstein]{hanson2025speedysplat}
Alex Hanson, Allen Tu, Geng Lin, Vasu Singla, Matthias Zwicker, and Tom
  Goldstein.
\newblock {Speedy-Splat}: Fast {3D Gaussian} splatting with sparse pixels and
  sparse primitives.
\newblock In \emph{CVPR}, pages 21537--21546, 2025.

\bibitem[Hedman et~al.(2018)Hedman, Philip, Price, Frahm, Drettakis, and
  Brostow]{hedman2018deep}
Peter Hedman, Julien Philip, True Price, Jan-Michael Frahm, George Drettakis,
  and Gabriel Brostow.
\newblock Deep blending for free-viewpoint image-based rendering.
\newblock \emph{ACM TOG}, 37\penalty0 (6), 2018.

\bibitem[Held et~al.(2025{\natexlab{a}})Held, Vandeghen, Deliege, Hamdi,
  Cioppa, Giancola, Vedaldi, Ghanem, Tagliasacchi, and
  Van~Droogenbroeck]{held2025trianglesplatting}
Jan Held, Renaud Vandeghen, Adrien Deliege, Abdullah Hamdi, Anthony Cioppa,
  Silvio Giancola, Andrea Vedaldi, Bernard Ghanem, Andrea Tagliasacchi, and
  Marc Van~Droogenbroeck.
\newblock Triangle splatting for real-time radiance field rendering,
  2025{\natexlab{a}}.

\bibitem[Held et~al.(2025{\natexlab{b}})Held, Vandeghen, Hamdi, Deliege,
  Cioppa, Giancola, Vedaldi, Ghanem, and
  Van~Droogenbroeck]{held2025convexsplatting}
Jan Held, Renaud Vandeghen, Abdullah Hamdi, Adrien Deliege, Anthony Cioppa,
  Silvio Giancola, Andrea Vedaldi, Bernard Ghanem, and Marc Van~Droogenbroeck.
\newblock {3D} convex splatting: Radiance field rendering with {3D} smooth
  convexes.
\newblock In \emph{CVPR}, pages 21360--21369, 2025{\natexlab{b}}.

\bibitem[Hijma et~al.(2023)Hijma, Heldens, Sclocco, van Werkhoven, and
  Bal]{hijma2023optimizing}
Pieter Hijma, Stijn Heldens, Alessio Sclocco, Ben van Werkhoven, and Henri~E.
  Bal.
\newblock Optimization techniques for gpu programming.
\newblock \emph{ACM Comput. Surv.}, 55\penalty0 (11), 2023.

\bibitem[Hou et~al.(2025)Hou, Rauwendaal, Li, Le, Farhadzadeh, Porikli, Bourd,
  and Said]{hou2025sortfree}
Qiqi Hou, Randall Rauwendaal, Zifeng Li, Hoang Le, Farzad Farhadzadeh, Fatih
  Porikli, Alexei Bourd, and Amir Said.
\newblock Sort-free {Gaussian} splatting via weighted sum rendering.
\newblock In \emph{ICLR}, 2025.

\bibitem[Huang et~al.(2024{\natexlab{a}})Huang, Yu, Chen, Geiger, and
  Gao]{huang20242d}
Binbin Huang, Zehao Yu, Anpei Chen, Andreas Geiger, and Shenghua Gao.
\newblock {2D Gaussian} splatting for geometrically accurate radiance fields.
\newblock In \emph{SIGGRAPH}, 2024{\natexlab{a}}.

\bibitem[Huang et~al.(2025)Huang, Subhajyoti~Mallick, Amat, Ruiz~Olle,
  Mosella-Montoro, Kerbl, Vicente~Carrasco, and De~la
  Torre]{huang2025echoesofcoliseum}
Junkai Huang, Saswat Subhajyoti~Mallick, Alejandro Amat, Marc Ruiz~Olle, Albert
  Mosella-Montoro, Bernhard Kerbl, Francisco Vicente~Carrasco, and Fernando
  De~la Torre.
\newblock Echoes of the coliseum: Towards 3d live streaming of sports events.
\newblock \emph{ACM TOG}, 44\penalty0 (4), 2025.

\bibitem[Huang et~al.(2024{\natexlab{b}})Huang, Bai, Guo, Li, and
  Guo]{huang2024erroranalysis}
Letian Huang, Jiayang Bai, Jie Guo, Yuanqi Li, and Yanwen Guo.
\newblock On the error analysis of {3D Gaussian} splatting and an optimal
  projection strategy.
\newblock In \emph{ECCV}, pages 247--263, 2024{\natexlab{b}}.

\bibitem[Höllein et~al.(2025)Höllein, Bo\v{z}i\v{c}, Zollhöfer, and
  Nie{\ss}ner]{hoellein20253dgslm}
Lukas Höllein, Alja\v{z} Bo\v{z}i\v{c}, Michael Zollhöfer, and Matthias
  Nie{\ss}ner.
\newblock {3DGS-LM}: Faster {Gaussian}-splatting optimization with
  {Levenberg-Marquardt}.
\newblock In \emph{ICCV}, 2025.

\bibitem[Jiang et~al.(2025)Jiang, Mao, Xu, Lu, Ren, Jin, Xu, Yu, Pang, Zhao,
  Lin, and Dai]{jiang2025anysplat}
Lihan Jiang, Yucheng Mao, Linning Xu, Tao Lu, Kerui Ren, Yichen Jin, Xudong Xu,
  Mulin Yu, Jiangmiao Pang, Feng Zhao, Dahua Lin, and Bo Dai.
\newblock {AnySplat}: Feed-forward {3D Gaussian} splatting from unconstrained
  views.
\newblock \emph{ACM TOG}, 44\penalty0 (6), 2025.

\bibitem[Kappel et~al.(2026)Kappel, Hahlbohm, and Scholz]{nerficg}
Moritz Kappel, Florian Hahlbohm, and Timon Scholz.
\newblock {NeRFICG}, 2026.

\bibitem[Kerbl et~al.(2023)Kerbl, Kopanas, Leimkuehler, and
  Drettakis]{kerbl3Dgaussians}
Bernhard Kerbl, Georgios Kopanas, Thomas Leimkuehler, and George Drettakis.
\newblock {3D Gaussian} splatting for real-time radiance field rendering.
\newblock \emph{ACM TOG}, 42\penalty0 (4), 2023.

\bibitem[Kerbl et~al.(2024)Kerbl, Meuleman, Kopanas, Wimmer, Lanvin, and
  Drettakis]{kerbl2024hierarchical}
Bernhard Kerbl, Andreas Meuleman, Georgios Kopanas, Michael Wimmer, Alexandre
  Lanvin, and George Drettakis.
\newblock A hierarchical {3D Gaussian} representation for real-time rendering
  of very large datasets.
\newblock \emph{ACM TOG}, 43\penalty0 (4), 2024.

\bibitem[Kheradmand et~al.(2024)Kheradmand, Rebain, Sharma, Sun, Tseng, Isack,
  Kar, Tagliasacchi, and Yi]{kheradmand243dgsmcmc}
Shakiba Kheradmand, Daniel Rebain, Gopal Sharma, Weiwei Sun, Yang-Che Tseng,
  Hossam Isack, Abhishek Kar, Andrea Tagliasacchi, and Kwang~Moo Yi.
\newblock {3D Gaussian} splatting as {Markov} chain {Monte Carlo}.
\newblock In \emph{NeurIPS}, 2024.

\bibitem[Kheradmand et~al.(2025)Kheradmand, Vicini, Kopanas, Lagun, Yi,
  Matthews, and Tagliasacchi]{kv2025stochasticsplats}
Shakiba Kheradmand, Delio Vicini, George Kopanas, Dmitry Lagun, Kwang~Moo Yi,
  Mark Matthews, and Andrea Tagliasacchi.
\newblock {StochasticSplats}: Stochastic rasterization for sorting-free {3D
  Gaussian} splatting.
\newblock In \emph{ICCV}, 2025.

\bibitem[Kingma and Ba(2015)]{kingma2014adam}
Diederik~P Kingma and Jimmy Ba.
\newblock Adam: A method for stochastic optimization.
\newblock In \emph{ICLR}, 2015.

\bibitem[Knapitsch et~al.(2017)Knapitsch, Park, Zhou, and
  Koltun]{Knapitsch2017}
Arno Knapitsch, Jaesik Park, Qian-Yi Zhou, and Vladlen Koltun.
\newblock {Tanks and Temples}: Benchmarking large-scale scene reconstruction.
\newblock \emph{ACM TOG}, 36\penalty0 (4), 2017.

\bibitem[Kotovenko et~al.(2025)Kotovenko, Grebenkova, and
  Ommer]{kotovenko2025edgs}
Dmytro Kotovenko, Olga Grebenkova, and Björn Ommer.
\newblock {EDGS}: Eliminating densification for efficient convergence of
  {3DGS}, 2025.

\bibitem[Kulhanek et~al.(2025)Kulhanek, Rakotosaona, Manhardt, Tsalicoglou,
  Niemeyer, Sattler, Peng, and Tombari]{kulhanek2025lodge}
Jonas Kulhanek, Marie-Julie Rakotosaona, Fabian Manhardt, Christina
  Tsalicoglou, Michael Niemeyer, Torsten Sattler, Songyou Peng, and Federico
  Tombari.
\newblock {LODGE}: Level-of-detail large-scale {Gaussian} splatting with
  efficient rendering.
\newblock In \emph{NeurIPS}, 2025.

\bibitem[Lan et~al.(2025)Lan, Shao, Lu, Zhang, Jiang, and
  Yang]{lan2025secondordergs}
Lei Lan, Tianjia Shao, Zixuan Lu, Yu Zhang, Chenfanfu Jiang, and Yin Yang.
\newblock {3DGS2}: Near second-order converging {3D Gaussian} splatting.
\newblock In \emph{SIGGRAPH}, 2025.

\bibitem[Li et~al.(2025)Li, Keller, Lin, and Khailany]{li2025gaurast}
Sixu Li, Ben Keller, Yingyan~(Celine) Lin, and Brucek Khailany.
\newblock {GauRast}: Enhancing gpu triangle rasterizers to accelerate {3D
  Gaussian} splatting.
\newblock In \emph{DAC}, 2025.

\bibitem[Li et~al.(2022)Li, Slavcheva, Zollhoefer, Green, Lassner, Kim,
  Schmidt, Lovegrove, Goesele, Newcombe, and Lv]{li2022neural3dv}
Tianye Li, Mira Slavcheva, Michael Zollhoefer, Simon Green, Christoph Lassner,
  Changil Kim, Tanner Schmidt, Steven Lovegrove, Michael Goesele, Richard
  Newcombe, and Zhaoyang Lv.
\newblock Neural {3D} video synthesis from multi-view video.
\newblock In \emph{CVPR}, pages 5511--5521, 2022.

\bibitem[Liao et~al.(2025)Liao, Ding, Cui, Gong, Hu, Wang, Li, Wang, Zhang, and
  Fu]{liao2025tensorcoregs}
Zimu Liao, Jifeng Ding, Siwei Cui, Ruixuan Gong, Boni Hu, Yi Wang, Hengjie Li,
  Hui Wang, Xingcheng Zhang, and Rong Fu.
\newblock {TC-GS}: A faster {Gaussian} splatting module utilizing tensor cores.
\newblock In \emph{SIGGRAPH Asia}, 2025.

\bibitem[Lin et~al.(2024)Lin, Li, Tang, Liu, Liu, Liu, Lu, Wu, Xu, Yan, and
  Yang]{lin2024vastgaussian}
Jiaqi Lin, Zhihao Li, Xiao Tang, Jianzhuang Liu, Shiyong Liu, Jiayue Liu,
  Yangdi Lu, Xiaofei Wu, Songcen Xu, Youliang Yan, and Wenming Yang.
\newblock {VastGaussian}: Vast {3D Gaussians} for large scene reconstruction.
\newblock In \emph{CVPR}, 2024.

\bibitem[Liu et~al.(2025)Liu, Sun, Chen, Wang, and Feng]{liu2025betasplatting}
Rong Liu, Dylan Sun, Meida Chen, Yue Wang, and Andrew Feng.
\newblock Deformable beta splatting.
\newblock In \emph{SIGGRAPH}, 2025.

\bibitem[Liu et~al.(2024)Liu, Zhan, Tang, Shan, Zeng, Lin, Liu, and
  Liu]{liu2024humangaussian}
Xian Liu, Xiaohang Zhan, Jiaxiang Tang, Ying Shan, Gang Zeng, Dahua Lin, Xihui
  Liu, and Ziwei Liu.
\newblock Humangaussian: Text-driven 3d human generation with gaussian
  splatting.
\newblock In \emph{CVPR}, pages 6646--6657, 2024.

\bibitem[Lu et~al.(2024)Lu, Yu, Xu, Xiangli, Wang, Lin, and
  Dai]{lu2024scaffoldgs}
Tao Lu, Mulin Yu, Linning Xu, Yuanbo Xiangli, Limin Wang, Dahua Lin, and Bo
  Dai.
\newblock {Scaffold-GS}: Structured {3D Gaussians} for view-adaptive rendering.
\newblock In \emph{CVPR}, 2024.

\bibitem[Luiten et~al.(2024)Luiten, Kopanas, Leibe, and
  Ramanan]{luiten2023dynamic}
Jonathon Luiten, Georgios Kopanas, Bastian Leibe, and Deva Ramanan.
\newblock {Dynamic 3D Gaussians}: Tracking by persistent dynamic view
  synthesis.
\newblock In \emph{3DV}, pages 800--809, 2024.

\bibitem[Luo et~al.(2025)Luo, Cui, Luo, Chu, and Li]{luo2025gaussianvrediting}
Zhaofeng Luo, Zhitong Cui, Shijian Luo, Mengyu Chu, and Minchen Li.
\newblock {VR-Doh}: Hands-on {3D} modeling in virtual reality.
\newblock \emph{ACM TOG}, 44\penalty0 (4), 2025.

\bibitem[Mai et~al.(2024)Mai, Hedman, Kopanas, Verbin, Futschik, Xu, Kuester,
  Barron, and Zhang]{mai2024ever}
Alexander Mai, Peter Hedman, George Kopanas, Dor Verbin, David Futschik,
  Qiangeng Xu, Falko Kuester, Jonathan~T. Barron, and Yinda Zhang.
\newblock {EVER}: Exact volumetric ellipsoid rendering for real-time view
  synthesis.
\newblock \emph{arXiv:2410.01804}, 2024.

\bibitem[Mallick et~al.(2024)Mallick, Goel, Kerbl, Steinberger, Carrasco, and
  De~La~Torre]{taming3dgs}
Saswat~Subhajyoti Mallick, Rahul Goel, Bernhard Kerbl, Markus Steinberger,
  Francisco~Vicente Carrasco, and Fernando De~La~Torre.
\newblock {Taming 3DGS}: High-quality radiance fields with limited resources.
\newblock In \emph{SIGGRAPH Asia}, 2024.

\bibitem[Matsuki et~al.(2024)Matsuki, Murai, Kelly, and
  Davison]{matsuki2024gaussian}
Hidenobu Matsuki, Riku Murai, Paul~HJ Kelly, and Andrew~J Davison.
\newblock Gaussian splatting slam.
\newblock In \emph{CVPR}, pages 18039--18048, 2024.

\bibitem[Mildenhall et~al.(2020)Mildenhall, Srinivasan, Tancik, Barron,
  Ramamoorthi, and Ng]{mildenhall2020nerf}
Ben Mildenhall, Pratul~P. Srinivasan, Matthew Tancik, Jonathan~T. Barron, Ravi
  Ramamoorthi, and Ren Ng.
\newblock {NeRF}: Representing scenes as neural radiance fields for view
  synthesis.
\newblock In \emph{ECCV}, pages 405--421, 2020.

\bibitem[Moenne-Loccoz et~al.(2024)Moenne-Loccoz, Mirzaei, Perel, de~Lutio,
  Martinez~Esturo, State, Fidler, Sharp, and Gojcic]{moenne2024gaussiantracer}
Nicolas Moenne-Loccoz, Ashkan Mirzaei, Or Perel, Riccardo de Lutio, Janick
  Martinez~Esturo, Gavriel State, Sanja Fidler, Nicholas Sharp, and Zan Gojcic.
\newblock {3D Gaussian Ray Tracing}: Fast tracing of particle scenes.
\newblock \emph{ACM TOG}, 43\penalty0 (6), 2024.

\bibitem[M\"{u}ller et~al.(2022)M\"{u}ller, Evans, Schied, and
  Keller]{mueller2022instant}
Thomas M\"{u}ller, Alex Evans, Christoph Schied, and Alexander Keller.
\newblock Instant neural graphics primitives with a multiresolution hash
  encoding.
\newblock \emph{ACM TOG}, 41\penalty0 (4), 2022.

\bibitem[Niemeyer et~al.(2025)Niemeyer, Manhardt, Rakotosaona, Oechsle,
  Duckworth, Gosula, Tateno, Bates, Kaeser, and Tombari]{niemeyer2024radsplat}
Michael Niemeyer, Fabian Manhardt, Marie-Julie Rakotosaona, Michael Oechsle,
  Daniel Duckworth, Rama Gosula, Keisuke Tateno, John Bates, Dominik Kaeser,
  and Federico Tombari.
\newblock {RadSplat}: Radiance field-informed {Gaussian} splatting for robust
  real-time rendering with 900+ fps.
\newblock In \emph{3DV}, pages 134--144, 2025.

\bibitem[Philip et~al.(2025)Philip, Ma, Clausen, Xian, Taşel, He, Yu, George,
  Yu, Pilarski, and Debevec]{philip2025detail}
Julien Philip, Li Ma, Pascal Clausen, Wenqi Xian, Ahmet~Levent Taşel, Mingming
  He, Xueming Yu, David~M. George, Ning Yu, Oliver Pilarski, and Paul Debevec.
\newblock Detail enhanced gaussian splatting for large-scale volumetric
  capture.
\newblock In \emph{SIGGRAPH Asia}, 2025.

\bibitem[Poirier-Ginter et~al.(2025)Poirier-Ginter, Hu, Lalonde, and
  Drettakis]{poirier2025editable}
Yohan Poirier-Ginter, Jeffrey Hu, Jean-Fran{\c{c}}ois Lalonde, and George
  Drettakis.
\newblock Editable physically-based reflections in raytraced gaussian radiance
  fields.
\newblock In \emph{SIGGRAPH Asia}, 2025.

\bibitem[Pumarola et~al.(2021)Pumarola, Corona, Pons-Moll, and
  Moreno-Noguer]{pumarola2021dnerf}
Albert Pumarola, Enric Corona, Gerard Pons-Moll, and Francesc Moreno-Noguer.
\newblock {D-NeRF}: Neural radiance fields for dynamic scenes.
\newblock In \emph{CVPR}, pages 10313--10322, 2021.

\bibitem[Radl et~al.(2024)Radl, Steiner, Parger, Weinrauch, Kerbl, and
  Steinberger]{radl2024stopthepop}
Lukas Radl, Michael Steiner, Mathias Parger, Alexander Weinrauch, Bernhard
  Kerbl, and Markus Steinberger.
\newblock {StopThePop}: Sorted {Gaussian} splatting for view-consistent
  real-time rendering.
\newblock \emph{ACM TOG}, 43\penalty0 (4), 2024.

\bibitem[Radl et~al.(2025)Radl, Windisch, Deixelberger, Hladky, Steiner,
  Schmalstieg, and Steinberger]{radl2025sof}
Lukas Radl, Felix Windisch, Thomas Deixelberger, Jozef Hladky, Michael Steiner,
  Dieter Schmalstieg, and Markus Steinberger.
\newblock {SOF}: Sorted opacity fields for fast unbounded surface
  reconstruction, 2025.

\bibitem[Ren et~al.(2025)Ren, Jiang, Lu, Yu, Xu, Ni, and Dai]{ren2025octreegs}
Kerui Ren, Lihan Jiang, Tao Lu, Mulin Yu, Linning Xu, Zhangkai Ni, and Bo Dai.
\newblock {Octree-GS}: Towards consistent real-time rendering with
  {LOD}-structured {3D Gaussians}.
\newblock \emph{IEEE TPAMI}, 2025.

\bibitem[Rota~Bul\`{o} et~al.(2024)Rota~Bul\`{o}, Porzi, and
  Kontschieder]{rotabulo2025revising}
Samuel Rota~Bul\`{o}, Lorenzo Porzi, and Peter Kontschieder.
\newblock Revising densification in {Gaussian} splatting.
\newblock In \emph{ECCV}, pages 347--362, 2024.

\bibitem[R\"{u}ckert et~al.(2022)R\"{u}ckert, Franke, and
  Stamminger]{ruckert2022adop}
Darius R\"{u}ckert, Linus Franke, and Marc Stamminger.
\newblock {ADOP}: Approximate differentiable one-pixel point rendering.
\newblock \emph{ACM TOG}, 41\penalty0 (4), 2022.

\bibitem[Sabour et~al.(2025)Sabour, Goli, Kopanas, Matthews, Lagun, Guibas,
  Jacobson, Fleet, and Tagliasacchi]{sabour2025spotlesssplats}
Sara Sabour, Lily Goli, George Kopanas, Mark Matthews, Dmitry Lagun, Leonidas
  Guibas, Alec Jacobson, David Fleet, and Andrea Tagliasacchi.
\newblock {SpotLessSplats}: Ignoring distractors in {3D Gaussian} splatting.
\newblock \emph{ACM TOG}, 44\penalty0 (2), 2025.

\bibitem[Sch\"{u}tz et~al.(2022)Sch\"{u}tz, Kerbl, and
  Wimmer]{schutz2022software}
Markus Sch\"{u}tz, Bernhard Kerbl, and Michael Wimmer.
\newblock Software rasterization of 2 billion points in real time.
\newblock \emph{PACMCGIT}, 5\penalty0 (3), 2022.

\bibitem[Schütz et~al.(2021)Schütz, Kerbl, and Wimmer]{SCHUETZ-2021-PCC}
Markus Schütz, Bernhard Kerbl, and Michael Wimmer.
\newblock Rendering point clouds with compute shaders and vertex order
  optimization.
\newblock \emph{CGF}, 40\penalty0 (4):\penalty0 115--126, 2021.

\bibitem[Schütz et~al.(2025)Schütz, Peters, Hahlbohm, Eisemann, Magnor, and
  Wimmer]{schuetz2025splatshop}
Markus Schütz, Christoph Peters, Florian Hahlbohm, Elmar Eisemann, Marcus
  Magnor, and Michael Wimmer.
\newblock Splatshop: Efficiently editing large {Gaussian} splat models.
\newblock \emph{CGF}, 2025.

\bibitem[Steiner et~al.(2025)Steiner, Köhler, Radl, Windisch, Schmalstieg, and
  Steinberger]{steiner2025aaags}
Michael Steiner, Thomas Köhler, Lukas Radl, Felix Windisch, Dieter
  Schmalstieg, and Markus Steinberger.
\newblock {AAA-Gaussians}: Anti-aliased and artifact-free {3D Gaussian}
  rendering.
\newblock In \emph{ICCV}, pages 27650--27659, 2025.

\bibitem[Studio(2025)]{lichtfeld2025}
LichtFeld Studio.
\newblock A high-performance c++ and cuda implementation of 3d gaussian
  splatting, 2025.

\bibitem[Szymanowicz et~al.(2024)Szymanowicz, Rupprecht, and
  Vedaldi]{szymanowicz2024splatterimage}
Stanislaw Szymanowicz, Christian Rupprecht, and Andrea Vedaldi.
\newblock {Splatter Image}: Ultra-fast single-view {3D} reconstruction.
\newblock In \emph{CVPR}, pages 10208--10217, 2024.

\bibitem[Talegaonkar et~al.(2025)Talegaonkar, Belhe, Ramamoorthi, and
  Antipa]{talegaonkar2025vol3dgs}
Chinmay Talegaonkar, Yash Belhe, Ravi Ramamoorthi, and Nicholas Antipa.
\newblock Volumetrically consistent {3D Gaussian} rasterization.
\newblock In \emph{CVPR}, 2025.

\bibitem[Tang et~al.(2024)Tang, Chen, Chen, Wang, Zeng, and Liu]{tang2024lgm}
Jiaxiang Tang, Zhaoxi Chen, Xiaokang Chen, Tengfei Wang, Gang Zeng, and Ziwei
  Liu.
\newblock {LGM}: Large multi-view {Gaussian} model for high-resolution {3D}
  content creation.
\newblock In \emph{ECCV}, pages 1--18, 2024.

\bibitem[Tosi et~al.(2024)Tosi, Zhang, Gong, Sandstr{\"o}m, Mattoccia, Oswald,
  and Poggi]{tosi2024nerfs}
Fabio Tosi, Youmin Zhang, Ziren Gong, Erik Sandstr{\"o}m, Stefano Mattoccia,
  Martin~R Oswald, and Matteo Poggi.
\newblock How nerfs and 3d gaussian splatting are reshaping slam: a survey.
\newblock \emph{arXiv preprint arXiv:2402.13255}, 4:\penalty0 1, 2024.

\bibitem[Tu et~al.(2025)Tu, Radl, Steiner, Steinberger, Kerbl, and de~la
  Torre]{tu2025vrsplat}
Xuechang Tu, Lukas Radl, Michael Steiner, Markus Steinberger, Bernhard Kerbl,
  and Fernando de~la Torre.
\newblock {VRSplat}: Fast and robust {Gaussian} splatting for virtual reality.
\newblock \emph{PACMCGIT}, 8\penalty0 (1), 2025.

\bibitem[von L{\"u}tzow and Nie{\ss}ner(2025)]{von2025linprim}
Nicolas von L{\"u}tzow and Matthias Nie{\ss}ner.
\newblock {LinPrim}: Linear primitives for differentiable volumetric rendering.
\newblock \emph{arXiv:2501.16312}, 2025.

\bibitem[Wang et~al.(2024)Wang, Yi, and Ma]{wang2024adrgaussian}
Xinzhe Wang, Ran Yi, and Lizhuang Ma.
\newblock {AdR-Gaussian}: Accelerating {Gaussian} splatting with adaptive
  radius.
\newblock In \emph{SIGGRAPH Asia}, 2024.

\bibitem[Wang et~al.(2025)Wang, Yang, Xu, Sun, Zhang, Chen, Bao, Peng, and
  Zhou]{wang2025freetimegs}
Yifan Wang, Peishan Yang, Zhen Xu, Jiaming Sun, Zhanhua Zhang, Yong Chen, Hujun
  Bao, Sida Peng, and Xiaowei Zhou.
\newblock {FreeTimeGS}: Free {Gaussian} primitives at anytime anywhere for
  dynamic scene reconstruction.
\newblock In \emph{CVPR}, pages 21750--21760, 2025.

\bibitem[Windisch et~al.(2025)Windisch, Köhler, Radl, Steiner, Schmalstieg,
  and Steinberger]{windisch2025lodofgaussians}
Felix Windisch, Thomas Köhler, Lukas Radl, Michael Steiner, Dieter
  Schmalstieg, and Markus Steinberger.
\newblock A {LoD} of {Gaussians}: Unified training and rendering for
  ultra-large scale reconstruction with external memory, 2025.

\bibitem[Wu et~al.(2024)Wu, Yi, Fang, Xie, Zhang, Wei, Liu, Tian, and
  Wang]{wu20244dgs}
Guanjun Wu, Taoran Yi, Jiemin Fang, Lingxi Xie, Xiaopeng Zhang, Wei Wei, Wenyu
  Liu, Qi Tian, and Xinggang Wang.
\newblock {4D Gaussian} splatting for real-time dynamic scene rendering.
\newblock In \emph{CVPR}, pages 20310--20320, 2024.

\bibitem[Wu et~al.(2025)Wu, Esturo, Mirzaei, Moenne-Loccoz, and
  Gojcic]{wu20253dgut}
Qi Wu, Janick~Martinez Esturo, Ashkan Mirzaei, Nicolas Moenne-Loccoz, and Zan
  Gojcic.
\newblock {3DGUT}: Enabling distorted cameras and secondary rays in {Gaussian}
  splatting.
\newblock In \emph{CVPR}, pages 26036--26046, 2025.

\bibitem[Xiang et~al.(2025)Xiang, Lv, Xu, Deng, Wang, Zhang, Chen, Tong, and
  Yang]{xiang2025structured}
Jianfeng Xiang, Zelong Lv, Sicheng Xu, Yu Deng, Ruicheng Wang, Bowen Zhang,
  Dong Chen, Xin Tong, and Jiaolong Yang.
\newblock Structured 3d latents for scalable and versatile 3d generation.
\newblock In \emph{CVPR}, pages 21469--21480, 2025.

\bibitem[Xu et~al.(2025)Xu, Peng, Wang, Blum, Barath, Geiger, and
  Pollefeys]{xu2025depthsplat}
Haofei Xu, Songyou Peng, Fangjinhua Wang, Hermann Blum, Daniel Barath, Andreas
  Geiger, and Marc Pollefeys.
\newblock {DepthSplat}: Connecting {Gaussian} splatting and depth.
\newblock In \emph{CVPR}, pages 16453--16463, 2025.

\bibitem[Xu et~al.(2024)Xu, Xu, Yu, Peng, Sun, Bao, and
  Zhou]{xu2024temporalgaussianhierarchy}
Zhen Xu, Yinghao Xu, Zhiyuan Yu, Sida Peng, Jiaming Sun, Hujun Bao, and Xiaowei
  Zhou.
\newblock Representing long volumetric video with temporal {Gaussian}
  hierarchy.
\newblock \emph{ACM TOG}, 43\penalty0 (6), 2024.

\bibitem[Yang et~al.(2025{\natexlab{a}})Yang, Wang, Lo, Zhang, Oruganti, and
  Kulkarni]{yang2025gsacc}
Mengtian Yang, Yipeng Wang, Chieh-Pu Lo, Xiuhao Zhang, Sirish Oruganti, and
  Jaydeep~P. Kulkarni.
\newblock {GSAcc}: Accelerate {3D Gaussian} splatting via depth speculation and
  gaussian-centric rasterization.
\newblock In \emph{DAC}, 2025{\natexlab{a}}.

\bibitem[Yang et~al.(2025{\natexlab{b}})Yang, Xu, Jiang, Lin, and
  Dai]{yang2025v3dg}
Xijie Yang, Linning Xu, Lihan Jiang, Dahua Lin, and Bo Dai.
\newblock Virtualized {3D Gaussians}: Flexible cluster-based level-of-detail
  system for real-time rendering of composed scenes.
\newblock In \emph{SIGGRAPH}. Association for Computing Machinery,
  2025{\natexlab{b}}.

\bibitem[Yang et~al.(2024{\natexlab{a}})Yang, Gao, Zhou, Jiao, Zhang, and
  Jin]{yang2024deformable3dgs}
Ziyi Yang, Xinyu Gao, Wen Zhou, Shaohui Jiao, Yuqing Zhang, and Xiaogang Jin.
\newblock Deformable {3D Gaussians} for high-fidelity monocular dynamic scene
  reconstruction.
\newblock In \emph{CVPR}, 2024{\natexlab{a}}.

\bibitem[Yang et~al.(2024{\natexlab{b}})Yang, Yang, Pan, and
  Zhang]{yang20244dgaussians}
Zeyu Yang, Hongye Yang, Zijie Pan, and Li Zhang.
\newblock Real-time photorealistic dynamic scene representation and rendering
  with {4D Gaussian} splatting.
\newblock In \emph{ICLR}, 2024{\natexlab{b}}.

\bibitem[Yariv et~al.(2023)Yariv, Hedman, Reiser, Verbin, Srinivasan, Szeliski,
  Barron, and Mildenhall]{yariv2023bakedsdf}
Lior Yariv, Peter Hedman, Christian Reiser, Dor Verbin, Pratul~P. Srinivasan,
  Richard Szeliski, Jonathan~T. Barron, and Ben Mildenhall.
\newblock {BakedSDF}: Meshing neural {SDFs} for real-time view synthesis.
\newblock In \emph{SIGGRAPH}, 2023.

\bibitem[Ye et~al.(2025{\natexlab{a}})Ye, Shao, and Zhou]{ye2025gaussiansurfel}
Keyang Ye, Tianjia Shao, and Kun Zhou.
\newblock When {Gaussian} meets surfel: Ultra-fast high-fidelity radiance field
  rendering.
\newblock \emph{ACM TOG}, 44\penalty0 (4), 2025{\natexlab{a}}.

\bibitem[Ye et~al.(2025{\natexlab{b}})Ye, Li, Kerr, Turkulainen, Yi, Pan,
  Seiskari, Ye, Hu, Tancik, and Kanazawa]{ye2025gsplat}
Vickie Ye, Ruilong Li, Justin Kerr, Matias Turkulainen, Brent Yi, Zhuoyang Pan,
  Otto Seiskari, Jianbo Ye, Jeffrey Hu, Matthew Tancik, and Angjoo Kanazawa.
\newblock gsplat: An open-source library for gaussian splatting.
\newblock \emph{Journal of Machine Learning Research}, 26\penalty0
  (34):\penalty0 1--17, 2025{\natexlab{b}}.

\bibitem[Ye et~al.(2024)Ye, Li, Liu, Qiao, and Dou]{ye2024absgs}
Zongxin Ye, Wenyu Li, Sidun Liu, Peng Qiao, and Yong Dou.
\newblock {AbsGS}: Recovering fine details for {3D Gaussia}n splatting, 2024.

\bibitem[Yu et~al.(2024{\natexlab{a}})Yu, Chen, Huang, Sattler, and
  Geiger]{yu2024mip}
Zehao Yu, Anpei Chen, Binbin Huang, Torsten Sattler, and Andreas Geiger.
\newblock {Mip-Splatting}: Alias-free {3D Gaussian} splatting.
\newblock In \emph{CVPR}, pages 19447--19456, 2024{\natexlab{a}}.

\bibitem[Yu et~al.(2024{\natexlab{b}})Yu, Sattler, and Geiger]{yu2024gof}
Zehao Yu, Torsten Sattler, and Andreas Geiger.
\newblock {Gaussian Opacity Fields}: Efficient adaptive surface reconstruction
  in unbounded scenes.
\newblock \emph{ACM TOG}, 43\penalty0 (6), 2024{\natexlab{b}}.

\bibitem[Zhang et~al.(2024)Zhang, Fang, Shrestha, Liang, Long, and
  Tan]{zhang2024radegs}
Baowen Zhang, Chuan Fang, Rakesh Shrestha, Yixun Liang, Xiaoxiao Long, and Ping
  Tan.
\newblock {RaDe-GS}: Rasterizing depth in {Gaussian} splatting, 2024.

\bibitem[Zhang et~al.(2018)Zhang, Isola, Efros, Shechtman, and
  Wang]{zhang2018lpips}
Richard Zhang, Phillip Isola, Alexei~A. Efros, Eli Shechtman, and Oliver Wang.
\newblock The unreasonable effectiveness of deep features as a perceptual
  metric.
\newblock In \emph{CVPR}, pages 586--595, 2018.

\bibitem[Zoomers et~al.(2025)Zoomers, Wijnants, Molenaers, Vanherck, Put,
  Jorissen, and Michiels]{zoomers2025progs}
Brent Zoomers, Maarten Wijnants, Ivan Molenaers, Joni Vanherck, Jeroen Put,
  Lode Jorissen, and Nick Michiels.
\newblock {PRoGS}: Progressive rendering of {Gaussian} splats.
\newblock In \emph{WACV}, 2025.

\bibitem[Zwicker et~al.(2001)Zwicker, Pfister, van Baar, and
  Gross]{zwicker2001ewasplat}
Matthias Zwicker, Hanspeter Pfister, Jeroen van Baar, and Markus Gross.
\newblock {EWA} volume splatting.
\newblock In \emph{Proc. of the Conference on Visualization}, pages 29--36,
  2001.

\end{thebibliography}
